\documentclass[lettersize,journal]{IEEEtran}

\usepackage[T1]{fontenc}
\usepackage{amsmath,amsfonts,amssymb,mathtools}
\usepackage{amsthm}
\usepackage{array}
\usepackage{textcomp}
\usepackage{stfloats}
\usepackage{url}
\usepackage{graphicx}
\usepackage{booktabs}
\usepackage{multirow}
\usepackage{xcolor}
\usepackage{microtype}
\usepackage[caption=false,font=normalsize,labelfont=sf,textfont=sf]{subfig}
\usepackage[linesnumbered,ruled,vlined]{algorithm2e}
\usepackage{cite}
\usepackage[hidelinks]{hyperref}
\usepackage[capitalize,noabbrev]{cleveref}
\newcommand{\methodname}{\texttt{PDA++}}

\theoremstyle{plain}

\theoremstyle{definition}

\theoremstyle{remark}

\begin{document}

% \title{PDA++: Field-Aligned Object Insertion in Remote Sensing Imagery}
% \title{PDA++: Field-Aligned Planning and Scene-Adaptive Generation for Remote Sensing Object Insertion}
\title{PDA++: Field-Aligned Planning and Scene-Adaptive Insertion in Remote Sensing}

\author{
    Xianchi~Dong,
    Yingyan~Hou,
    Chao~Ren,~\IEEEmembership{Member, IEEE},
    Wanxuan~Lu,
    Zihan~Wei,
    Hongfeng~Yu,
    Yixiao~Wang,~\IEEEmembership{Member, IEEE},
    Chubo~Deng,
    and Xian~Sun$^*$,~\IEEEmembership{Senior Member, IEEE}%
% \thanks{$^*$Corresponding author: Xian Sun.}%
% \thanks{Xianchi~Dong, Xian Sun, Yingyan~Hou and Zihan~Wei are with the Aerospace Information Research Institute, Chinese Academy of Sciences, Beijing 100094, China, the University of Chinese Academy of Sciences, Beijing 100190, China, the School of Electronic, Electrical and Communication Engineering, University of Chinese Academy of Sciences, Beijing 100190, China, and also the Key Laboratory of Target Cognition and Application Technology (TCAT), Aerospace Information Research Institute, Chinese Academy of Sciences, Beijing 100094, China (e-mail: dongxianchi24@mails.ucas.ac.cn; sunxian@aircas.ac.cn; houyy@aircas.ac.cn; weizihan24@mails.ucas.ac.cn).}%
% \thanks{Chao~Ren, Wanxuan~Lu, Hongfeng~Yu, Yixiao~Wang and Chubo~Deng are with the Aerospace Information Research Institute, Chinese Academy of Sciences, Beijing 100094, China, and also the Key Laboratory of Target Cognition and Application Technology (TCAT), Aerospace Information Research Institute, Chinese Academy of Sciences, Beijing 100094, China (e-mail: renc0003@e.ntu.edu.sg; luwx@aircas.ac.cn; dengcb@aircas.ac.cn; yuhf@aircas.ac.cn; wangyixiao@aircas.ac.cn; dengcb@aircas.ac.cn).}%
% \thanks{This work was supported by the National Natural Science Foundation of China under Grant 62425115.}%
}

% The paper headers
% \markboth{IEEE Transactions on Pattern Analysis and Machine Intelligence,~Vol.~XX, No.~XX, Month~2026}%
% \markboth{journal of xxxx}
% {HOU \MakeLowercase{\textit{et al.}}: Plan, Decouple, Assimilate: Environment-Adaptive Object Insertion in Remote Sensing Imagery}

\maketitle

\begin{abstract}
Remote sensing recognition is often constrained by scarce observations of rare targets and costly annotations, making realistic synthetic augmentation particularly valuable for few-shot and long-tailed scenarios. Object insertion provides an efficient way to increase target diversity while preserving authentic background scenes, but realistic insertion in overhead imagery requires the generated target to adapt coherently to its surrounding environment. To this end, we propose \methodname{}, a unified environment-aware object insertion framework organized as \underline{P}lan, \underline{D}ecouple, and \underline{A}ssimilate. Planning determines scene-compatible poses through an affordance field that combines geometric clearance with structure- and scale-aware cues. Decoupling introduces a pose-conditioned background that provides precise spatial guidance together with target-scene context, allowing the reference object to preserve its identity while adapting to the target observation. This construction also naturally provides pixel-level masks for segmentation augmentation. Assimilation further improves local coherence by aligning multi-scale texture distributions through optimal transport. On the optical benchmark, \methodname{} achieves a whole-image FID of 6.28 and improves average few-shot recognition mAP50 by 17.69 points, corresponding to a 28.8\% relative gain over the real-data baseline. On SAR imagery, it improves ship detection by 4.10 mAP50 points and remains effective under cross-dataset transfer and amorphous-target insertion. Code is available at \url{https://github.com/lisheyu972/PDA_PLUS}.
\end{abstract}

% \newpage

\begin{figure*}[!t] 
\centering 
\includegraphics[width=0.95\textwidth]{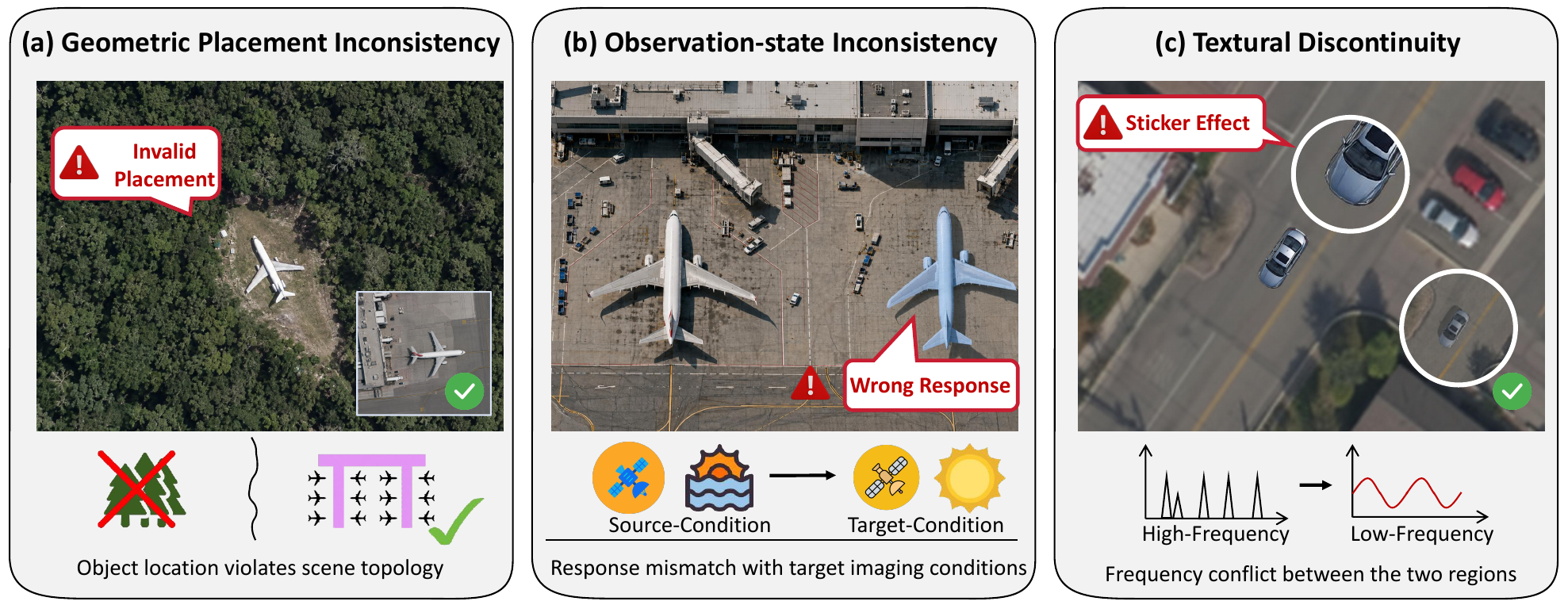} 
\caption{Illustration of three critical issues in remote sensing object insertion. 
(a) Geometric placement inconsistency, where objects are inserted into spatially or semantically implausible regions; 
(b) observation-state inconsistency, where the inserted target retains sensing and acquisition characteristics that are incompatible with the target scene, resulting in mismatched sensor response, illumination, or acquisition geometry; 
(c) textural discontinuity, where the roughness and texture statistics of the inserted target are incompatible with the surrounding background.}

\vspace{-6pt}
\label{fig:challenge} 
\end{figure*}

\section{Introduction}

\IEEEPARstart{R}{emote} sensing object detection and recognition play an important role in Earth observation applications, particularly in large-scale
environmental monitoring and the analysis of critical infrastructure~\cite{rekavandi2025guide,zhou2021aircraft,bandarupally2020detection}.
Their performance, however, depends heavily on the availability of large-scale annotated data~\cite{gui2024remote,li2020object,ding2021object}. In practice,
observations of many important targets remain sparse because of limited revisit opportunities and practical acquisition constraints. Rare and strategically 
important targets therefore tend to exhibit pronounced long-tailed distributions~\cite{gao2024yolo,wang2022remote}. Moreover, the efficacy of deep detectors exhibits notable vulnerability to data-induced perturbations~\cite{mei2024comprehensive}. The problem is further complicated by substantial appearance variation across scenes and imaging conditions, which makes sample diversity as important as data quantity. Collecting additional real observations is often difficult to control and may still fail to cover rare configurations of interest. Synthetic augmentation offers an alternative by increasing the occurrence and diversity of such targets without requiring new acquisition campaigns. Its usefulness, however, depends strongly on whether the generated samples remain compatible with the physical and observational characteristics of real remote sensing imagery. Realistic augmentation is consequently particularly valuable for improving recognition under few-shot and long-tailed settings~\cite{pan2025earthsynth,tang2025terragen,yang2025task}.

Object insertion provides a flexible way to increase target diversity while retaining authentic remote sensing backgrounds. Conventional copy-based augmentation is simple and computationally efficient, but the pasted object remains largely unchanged after being transferred to a new scene and often appears incompatible with its surroundings. Such inconsistencies are especially problematic for downstream augmentation because the synthesized target is not merely required to look plausible in isolation. It must also provide supervision that follows the visual statistics and spatial structure of the destination scene. Recent diffusion-based editing methods, including AnyDoor~\cite{chen2024anydoor}, MimicBrush~\cite{chen2024zero}, UniCombine~\cite{wang2025unicombine},  Insert Anything~\cite{song2025insert}, Qwen Image Edit~\cite{wu2025qwen}, and OminiControl~\cite{tan2025ominicontrol}, have substantially improved object-level generation in natural imagery. Their ability to synthesize missing content from reference conditions suggests a promising direction for remote sensing augmentation. Nevertheless, remote sensing imagery differs fundamentally from natural imagery because object appearance depends strongly on the observation process in addition to scene geometry. The same semantic target can exhibit markedly different responses when observed under different sensing conditions. Directly transferring a reference appearance to a new scene can therefore produce an object that is recognizable yet inconsistent with the destination observation. This makes remote sensing insertion more than a spatial editing problem 
and requires the generated content to adapt to the scene in which it is placed.

As illustrated in Fig.~\ref{fig:challenge}, this incompatibility can emerge throughout the insertion process. An object may already be implausible when
its location or pose conflicts with the underlying scene structure, such as an aircraft synthesized outside an admissible airport region or a ship placed on land. Even when the placement is valid, the inserted target may retain characteristics inherited from its source observation and therefore appear inconsistent with the imaging state of the destination scene. Such discrepancies can manifest differently in optical and SAR imagery because the rendered appearance depends strongly on the sensing modality. Residual artifacts may also remain around the insertion boundary when local texture statistics are not compatible with the surrounding background. These failure modes are closely coupled because the selected pose determines the environmental context available to subsequent generation. An inappropriate placement cannot be corrected solely through appearance synthesis, while accurate placement does not guarantee that the rendered target will match its surroundings. Reliable insertion therefore requires the synthesized content to adapt progressively to its destination environment rather than treating placement and generation as independent operations.

Motivated by this observation, we propose \methodname{}, a unified Plan Decouple Assimilate framework that progressively adapts inserted content to the target scene. Planning determines a scene-compatible pose through an Affordance Field that combines geometric clearance with structure-aware and scale-aware evidence. The resulting pose defines not only where the target should appear but also the local context used during generation. Decoupling then introduces a pose-conditioned background that communicates precise spatial support together with the target observation context. The reference object can therefore retain its semantic identity while its rendered appearance is adapted to the destination scene. This construction also naturally associates each synthesized target with a pixel-level mask and thereby extends synthetic augmentation from recognition to segmentation.
Assimilation acts during generation to reduce residual local discrepancy. It represents the inserted region and its neighborhood through multi-scale texture statistics and aligns their distributions with optimal transport. The resulting process couples scene-aware placement with environment-adaptive generation so that the output is determined jointly by the reference identity and the destination observation.

A preliminary version of this work appeared at ICML 2026~\cite{hou2026plan}, where the Plan Decouple Assimilate framework was
introduced for remote sensing object insertion. This journal version substantially extends the conference work through a redesign of both planning and generation. The original clearance-based planner is upgraded to an Affordance Field that better accounts for scene structure and object scale. The generation stage is reformulated around pose-conditioned scene context, while the original texture objective is replaced by multi-scale optimal-transport alignment. These changes replace the conference ASA and NATA designs with a more unified environment-adaptive generation process. The journal version also extends the functionality of the framework beyond the original box-level augmentation setting. Pixel-level masks produced by the pose-conditioned construction enable segmentation augmentation without additional manual annotation, and recursive synthesis allows several targets
to be generated within the same scene while preserving compatibility with previous insertions. We further conduct controlled component analyses to separate the effect of the redesigned modules from that of the updated generative backbone. Together, these extensions broaden the use of the original framework while preserving its applicability across the optical and SAR settings established in the conference version.

The main contributions of this work are summarized as follows:
\begin{itemize}

    \item We introduce a unified environment-aware formulation for remote sensing object insertion, where the inserted target is progressively adapted to the target scene from placement to appearance. This perspective links scene-compatible layout with observation-aware generation and local harmonization in a single insertion pipeline.

    \item We propose \methodname{}, which improves the original Plan--Decouple--Assimilate framework through an Affordance Field planner, pose-conditioned scene representation, and manifold-aligned texture guidance. These designs substantially improve insertion fidelity, achieving a whole-image FID of 6.28 on the optical benchmark while also enabling pixel-level supervision for segmentation augmentation.

    \item We validate \methodname{} extensively on both optical and SAR imagery under diverse augmentation settings. On MAR20-11-FewShot, the generated samples improve average recognition mAP50 by 17.69 points, corresponding to a 28.8\% relative gain over the real-data baseline. On SAR ship detection, the improvement reaches 4.10 mAP50 points, with consistent effectiveness under cross-dataset transfer and amorphous-target insertion.

\end{itemize}

\section{Related Work}

\paragraph{Generative Models}
Generative models have undergone a major paradigm shift from generative adversarial networks (GANs)~\cite{goodfellow2014generative} to denoising diffusion probabilistic models~\cite{ho2020denoising}. GANs once dominated image generation because of their strong visual fidelity and efficient sampling, but adversarial optimization is often unstable and may limit distribution coverage. Diffusion models instead formulate generation as the reverse of a stochastic noising process, offering improved training stability and sample diversity~\cite{dhariwal2021diffusion}. Their strong performance in high-fidelity synthesis has made them a dominant paradigm in modern generative modeling. Recent work further extends diffusion models toward unified generation-and-editing frameworks that support a broad range of conditional synthesis tasks within a shared backbone~\cite{fu2025univg}. Latent diffusion models (LDMs)~\cite{rombach2022high} improve efficiency by moving the generative process into a lower-dimensional latent space while largely preserving visual quality.

Parameter-efficient adaptation further facilitates the transfer of pretrained generative models to specialized domains. Low-Rank Adaptation (LoRA)~\cite{hu2022lora} introduces only a small set of trainable parameters and therefore avoids expensive full-model optimization. Together with latent diffusion and flexible conditioning mechanisms, these developments provide the technical basis for modern object insertion frameworks.

\paragraph{Object Insertion and Image Editing}
Object insertion aims to integrate a foreground object into a new background while maintaining visual plausibility and semantic coherence. It can be viewed as a specialized form of image composition whose realism depends on whether the inserted content is compatible with the geometry and appearance of the destination scene~\cite{niu2021making}. Conventional copy-based or blending methods are computationally efficient, but they largely preserve the source appearance and provide little capability to adapt the inserted object to its new environment.

Recent diffusion-based editing models have substantially improved this capability. Representative methods such as AnyDoor~\cite{chen2024anydoor}, MimicBrush~\cite{chen2024zero}, Insert Anything~\cite{song2025insert}, UniCombine~\cite{wang2025unicombine}, Qwen Image Edit~\cite{wu2025qwen}, OminiControl~\cite{tan2025ominicontrol}, and AnyEdit~\cite{yu2025anyedit} demonstrate increasingly flexible reference-guided object manipulation. Controllable diffusion models can further exploit visual or spatial conditions to constrain the generation process~\cite{zhang2023adding}. Of particular relevance to our work, OminiControl~\cite{tan2025ominicontrol} and Insert Anything~\cite{song2025insert} directly place reference information into the generation context rather than relying on a dedicated reference encoder. This in-context formulation allows subject identity to interact with target-scene information through the native attention mechanism.

Recent studies have also explored more specialized composition settings. Zero-shot methods investigate how intrinsic scene cues can guide object integration~\cite{zhang2025zerocomp}, while feature-level approaches improve the interaction between reference content and target appearance~\cite{li2025aicomposer}. Personalized insertion and 3D scene editing further extend the controllability of object composition~\cite{zhang2025freeinsert,li2025freeinsert}. Meanwhile, HiddenObjects develops scalable spatial priors for object placement~\cite{schouten2026hiddenobjects}, and Region-to-Region improves local harmonization through region-aware injection~\cite{zhang2025region}. Physical compatibility has also received increasing attention: SpotLight addresses scene-aware relighting~\cite{fortier2024spotlight}, whereas SSN explicitly models soft shadows during composition~\cite{sheng2021ssn}.

Despite these advances, most existing insertion models are developed primarily for natural imagery and datasets such as COCO~\cite{lin2014microsoft}. Their direct application to remote sensing imagery therefore suffers from a substantial domain gap~\cite{liu2024diffusion}. Unlike perspective natural scenes, remote sensing observations exhibit stronger topological constraints and sensing-dependent appearance variations. A target that appears plausible in isolation may consequently remain incompatible with its destination observation. This difference motivates insertion frameworks that account for both scene structure and the imaging characteristics of remote sensing data.

\paragraph{Environment-Aware Generation and In-Context Conditioning}
A central challenge in conditional generation is to make synthesized content respond coherently to the environment in which it appears. Beyond preserving object identity, realistic synthesis requires the generated appearance to remain compatible with the spatial layout and observation characteristics of the surrounding scene. Recent studies suggest that large-scale diffusion models already encode substantial environmental knowledge. DiffusionLight~\cite{phongthawee2024diffusionlight}, for example, demonstrates that scene illumination can be recovered from pretrained generative representations. Related studies show that diffusion features retain geometric information useful for monocular depth estimation~\cite{ke2024repurposing}, while RGBX~\cite{zeng2024rgbx} reveals material- and lighting-aware representations that support intrinsic decomposition and relighting. These findings indicate that pretrained generative models contain useful priors for adapting synthesized content to its environment.

In-context conditioning provides a direct way to expose such environmental evidence during generation. OminiControl~\cite{tan2025ominicontrol} and Insert Anything~\cite{song2025insert} place reference and target information within a shared attention context, allowing subject identity to interact directly with the destination scene. This mechanism is particularly suitable for object insertion because the reference content must remain recognizable while its appearance changes with the new environment. Existing studies, however, largely focus on natural imagery, where environmental variation is dominated by visible-scene factors such as illumination and viewpoint. Remote sensing observations additionally involve modality-dependent responses and acquisition geometry, making environment adaptation more closely tied to the underlying imaging process.

\paragraph{Generative Models in Remote Sensing}
Generative modeling has also become increasingly important in remote sensing. Diffusion-based methods have been applied to image restoration, including remote sensing super-resolution~\cite{liu2022diffusion,wang2025semantic}. Their use for data augmentation has expanded as well, providing an alternative means of increasing training diversity when real observations are limited~\cite{sousa2025data,yuan2023efficient}. More recent studies investigate controllable remote sensing synthesis through image editing~\cite{zhenyuan2026rsedit}, change-oriented generation~\cite{tang2024changeanywhere}, and task-specific synthetic data construction~\cite{martin2026generating}.

Foundation-style generative models further broaden the scale and controllability of remote sensing synthesis. DiffusionSat~\cite{khanna2024diffusionsat} incorporates geospatial metadata into conditional satellite image generation, while CRS-Diff~\cite{tang2024crs} exploits geospatial conditions for controllable synthesis. MetaEarth~\cite{yu2025metaearth} and Text2Earth~\cite{liu2025text2earth} extend generation toward global-scale and text-driven settings, with EcoMapper~\cite{goktepe2025ecomapper} further introducing climate-aware environmental information. Other approaches use segmentation or land-cover information as spatial guidance~\cite{toker2024satsynth,deng2025synthesizing}. Large geospatial models such as CrossEarth~\cite{gong2025crossearth} and SARATR-X~\cite{li2025saratr} also highlight the importance of modality-aware pretraining. For SAR generation in particular, recent work has begun to incorporate physical or geometric priors into diffusion-based synthesis~\cite{zhang2026geodiff,debuysere2025quantitative}.

Despite this progress, high-fidelity object insertion remains comparatively underexplored in remote sensing~\cite{han2025exploring}. Existing methods generally address broad image generation or isolated aspects of image editing, whereas insertion requires the generated target to remain compatible with the destination environment throughout the synthesis process. A plausible pose alone is insufficient if the resulting object retains an incompatible observation state, while visually coherent generation can still expose artifacts around the insertion boundary. Recent studies on shadow consistency and texture harmonization further illustrate the importance of environment-aware integration for synthetic imagery~\cite{zhang2025shadow,tsai2017deep,cong2020dovenet}.

Our work addresses this gap by viewing remote sensing insertion as a progressive environment-adaptation process. Instead of treating object insertion as generic image editing, the proposed framework first establishes scene-compatible spatial support and then adapts the generated content to the destination observation. Residual local discrepancies are subsequently reconciled through texture-aware guidance, allowing placement and appearance adaptation to operate within a unified generation pipeline.

\section{Preliminaries}

\subsection{Problem Formulation}

Given a remote sensing background image $I_{bg}\in\mathbb{R}^{H\times W\times3}$ and a target object image $I_{sub}\in\mathbb{R}^{H\times W\times3}$,
our goal is to generate a composite image $I_{final}\in\mathbb{R}^{H\times W\times3}$ in which the target is inserted at a scene-compatible location.
The insertion process couples scene-aware planning with conditional generation. Planning determines an insertion mask
$M\in\{0,1\}^{H\times W}$ whose spatial support is compatible with the background scene. Given the resulting $M$, the background $I_{bg}$, and the
reference object $I_{sub}$, the generation process synthesizes the target within the planned region while adapting its appearance to the destination
observation. The resulting composite is modeled as
\begin{equation}
    I_{final}
    \sim
    p_{\theta}
    \left(
        \cdot
        \mid
        I_{bg},
        M,
        I_{sub}
    \right).
    \label{eq:problem_formulation}
\end{equation}

\subsection{Subject-Driven Condition Injection}
\label{sec:Subject-Driven Condition Injection}

To integrate the subject reference into the DiT architecture, we follow the in-context conditioning paradigm of OminiControl~\cite{tan2025ominicontrol}.
Latent tokens derived from $I_{sub}$ are concatenated with the noisy image tokens to form a joint sequence, allowing the native multimodal attention to
model their interaction without an auxiliary reference encoder. Since subject-driven generation requires identity preservation rather than strict
spatial correspondence, we adopt shifted positional encoding for the reference tokens. Their position indices are translated by a fixed offset so that the
reference and generation tokens occupy disjoint coordinate ranges in the rotary embedding space. This separation weakens direct local correspondence
between the two token groups and encourages the attention mechanism to capture subject-level semantic information.

The resulting in-context representation provides the identity condition used throughout our framework. During generation, the reference tokens preserve the
semantic identity of $I_{sub}$, while the target background supplies the scene context required to adapt the inserted object to the destination observation,
as detailed in Section~\ref{decouple}.

\section{Methodology}
\subsection{Overview}
\label{sec:method}
Remote sensing object insertion requires the synthesized target to remain compatible with its destination scene throughout both placement and
generation. A plausible location alone is insufficient if the generated appearance remains inconsistent with the target observation, while visually
coherent generation can still exhibit local artifacts when the inserted content does not match its surroundings. To address these coupled
requirements, \methodname{} organizes object insertion into two stages. Stage I performs scene-aware planning, and Stage II adapts the inserted
content to the target observation through Decoupling and Assimilation. The overall procedure is summarized as

\begin{equation}
\begin{aligned}
\mathbf{p}^{*}
&=
\arg\max_{\mathbf{x},\theta}
\mathcal{A}(\mathbf{x},\theta;c),
\\
\tilde{I}_{bg}
&=
\mathcal{T}(I_{bg},M,M_{seg}),
\\
\hat{v}_{t}
&=
v_{\theta}(x_t,t)
-
\lambda(t)
\nabla_{x_t}
\mathcal{L}_{tex}(x_t).
\end{aligned}
\label{eq:overview}
\end{equation}

The first line describes Planning, where the affordance field $\mathcal{A}$ determines the object pose $\mathbf{p}^{*}$. The planned pose
defines the insertion region $M$ and is used to transform the subject mask into $M_{seg}$. The second line constructs the pose-conditioned background
$\tilde{I}_{bg}$, which communicates the planned spatial support while retaining the surrounding scene as observation context. The final line
describes texture-guided generation. The flow velocity $\mathbf{v}_{\theta}$ is corrected by the gradient of $\mathcal{L}_{\mathrm{tex}}$, 
whose statistics are computed from the inserted region and its local neighborhood $M_{env}$. The following sections describe each component in detail.

\textbf{Stage I (Planning).}
Planning determines where the target can be inserted and how it should be oriented in the scene. The proposed Affordance Field extends the
clearance-based criterion of the conference version by incorporating scene structure and object-scale compatibility. The resulting pose
$\mathbf{p}^{*}$ provides the spatial constraint used by the subsequent generation stage. Details are provided in Section~\ref{plan}.

\begin{figure*}[!t]
\centering
\includegraphics[width=\textwidth]{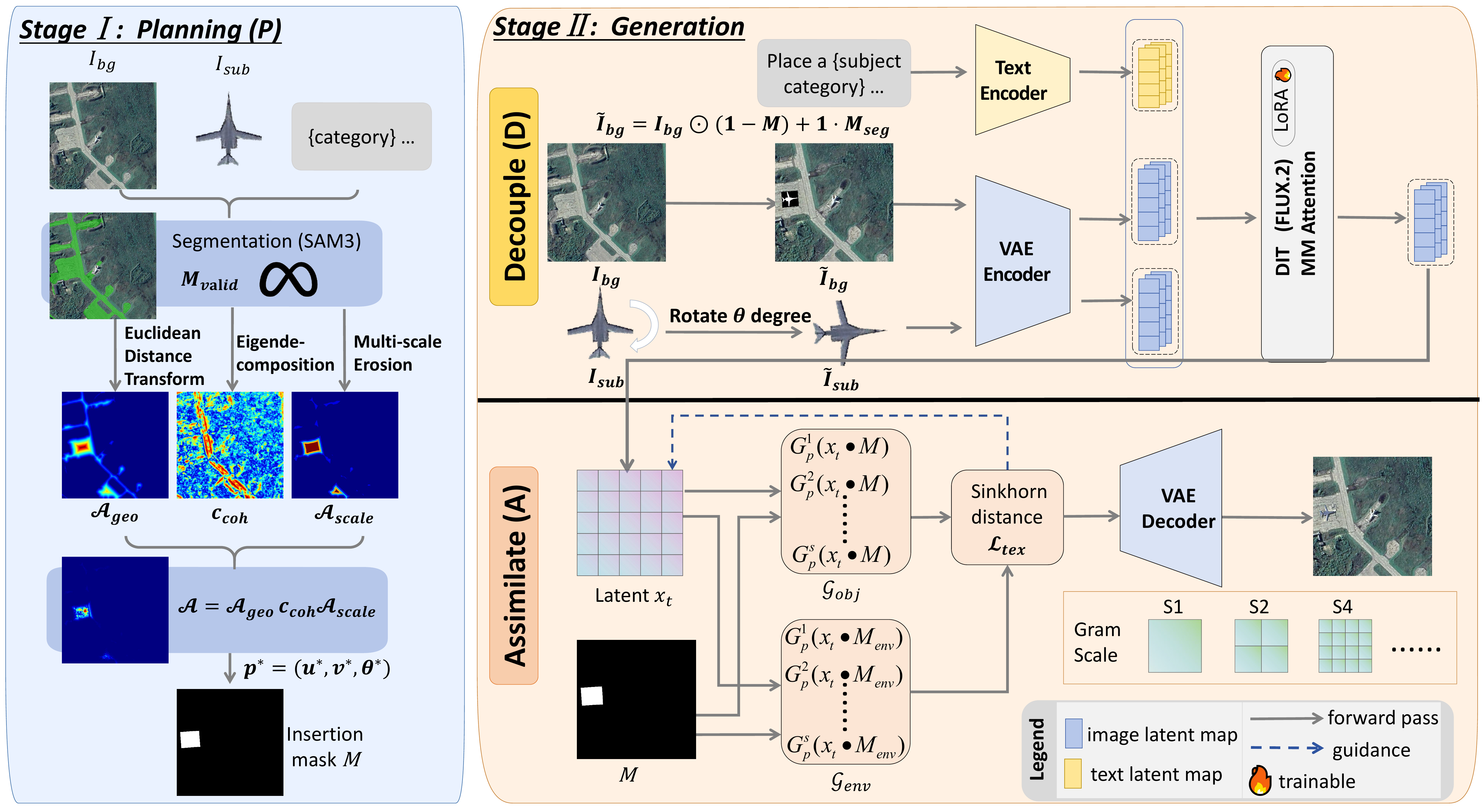}
\caption{Overall pipeline of \methodname{}. Stage I performs scene-aware
Planning (\texttt{P}) to determine the insertion pose. Stage II performs
Decoupling (\texttt{D}) and Assimilation (\texttt{A}) to adapt the inserted
target to the destination observation and improve its local compatibility
with the surrounding scene.}
\label{fig:2}
\vspace{-6pt}
\end{figure*}

\textbf{Stage II (Generation).}
Generation takes the planned pose together with the reference subject and target background as conditions. Decoupling converts the planned geometry
into pixel-level guidance through the pose-conditioned background $\tilde{I}_{bg}$. This representation preserves the surrounding observation
context, allowing the reference identity to be retained while its appearance is adapted to the destination scene. The same construction also provides
the object mask $M_{seg}$ for downstream segmentation augmentation. Assimilation subsequently acts on the sampling trajectory through
$\mathcal{L}_{\mathrm{tex}}$, which compares local texture statistics within $M$ and $M_{env}$ and corrects residual appearance discrepancies. Details
of these two components are given in Sections~\ref{decouple} and~\ref{ass}, respectively.

The two stages are modular. Planning supplies the geometry required for generation, whereas the generation process does not depend on how the
candidate pose is obtained. This separation allows the same generation formulation to operate with different planning strategies and across
different remote sensing modalities.

\begin{algorithm}[t]
\small
\caption{The proposed \methodname{} method}
\label{alg:pda_pipeline}
\SetAlgoNoLine
\LinesNumbered
\DontPrintSemicolon

\KwIn{
Background $I_{bg}$,
subject $I_{sub}$ with mask $M_{sub}$,
category set $\mathcal{C}_{category}$,
class $c$
}
\KwOut{Composite image $I_{final}$}

\tcp*[h]{\textbf{Stage I  Planning}}

$M_{valid}
\leftarrow
\mathrm{SemanticParse}(I_{bg},\mathcal{C}_{category})$\;

$\mathcal{A}(\mathbf{x},\theta;c)
\leftarrow
\mathcal{A}_{\mathrm{geo}}(\mathbf{x})
\mathcal{A}_{\mathrm{struct}}(\mathbf{x},\theta)
\mathcal{A}_{\mathrm{scale}}(\mathbf{x};c)$\;

$\mathbf{p}^{*}
\leftarrow
\arg\max_{\mathbf{x},\theta}
\mathcal{A}(\mathbf{x},\theta;c)$\;

$M
\leftarrow
\mathrm{BoxMask}(\mathbf{p}^{*})$,
\quad
$M_{seg}
\leftarrow
\mathrm{Transform}(M_{sub},\mathbf{p}^{*})$\;

$M_{env}
\leftarrow
\mathrm{Dilate}(M)\setminus M$\;

\tcp*[h]{\textbf{Stage II  Generation}}

$\tilde{I}_{bg}
\leftarrow
\mathcal{T}(I_{bg},M,M_{seg})$\;

$z_{bg}
\leftarrow
\mathrm{VAE.Enc}(\tilde{I}_{bg})$,
\quad
$z_{sub}
\leftarrow
\mathrm{VAE.Enc}(I_{sub})$\;

$T\leftarrow50$,
\quad
$1=t_0>t_1>\cdots>t_T=0$\;

$x_{t_0}
\sim
\mathcal{N}(0,\mathbf{I})$\;

\For{$i=0,1,\ldots,T-1$}{
    $v_{\theta}
    \leftarrow
    \mathrm{DiT}
    (x_{t_i},t_i,z_{bg},z_{sub})$\;

    $\mathcal{G}_{obj}
    \leftarrow
    \mathcal{G}(x_{t_i};M)$,
    \quad
    $\mathcal{G}_{env}
    \leftarrow
    \mathcal{G}(x_{t_i};M_{env})$\;

    $\mathcal{L}_{tex}
    \leftarrow
    \mathcal{W}_{\varepsilon}
    (\mathcal{G}_{obj},\mathcal{G}_{env})$\;

    $\hat{v}_{t_i}
    \leftarrow
    v_{\theta}
    -
    \lambda(t_i)
    \nabla_{x_{t_i}}\mathcal{L}_{tex}$\;

    $x_{t_{i+1}}
    \leftarrow
    \mathrm{ODESolver}
    (x_{t_i},\hat{v}_{t_i},t_i,t_{i+1})$\;
}

$I_{final}
\leftarrow
\mathrm{VAE.Dec}(x_{t_T})$\;

\Return{$I_{final}$}

\end{algorithm}

\subsection{Affordance-Aware Scene Layout Planner ({\tt{P}})}
\label{plan}

\paragraph{Semantic-Geometric Parsing}
We first identify the functional regions of the scene. SegEarthOV3~\cite{li2025segearth}, an open-vocabulary segmentation model based on SAM3~\cite{carion2025sam}, is applied to the background $I_{bg}$. Given a target-specific category set $\mathcal{C}_{category}$, the segmentation map $S$ is converted into a binary valid-region mask $M_{valid}\in\{0,1\}^{H\times W}$:
\begin{equation}
    M_{valid}(u, v) =
    \begin{cases}
    1 & \text{if } S(u, v) \in \mathcal{C}_{category}, \\
    0 & \text{otherwise}.
    \end{cases}
\end{equation}
Here, $(u,v)$ indexes a pixel and $S(u,v)$ denotes its predicted semantic class. The resulting mask restricts object placement to semantically permissible regions.

\paragraph{Affordance Field Formulation}
The conference version determines the insertion pose using clearance alone. It places the object near the maximum of the distance field and derives its orientation from the corresponding field gradient. This criterion does not account for whether the available spatial support is appropriate for the target scale. Moreover, the distance gradient mainly reflects boundary geometry and becomes unstable near the clearance maximum. We therefore introduce an \emph{affordance field} $\mathcal{A}(\mathbf{x},\theta;c)$ to evaluate the suitability of a candidate pose. For a position $\mathbf{x}=(u,v)\in M_{valid}$ and orientation $\theta$, the field is conditioned on target class $c$ and defined as
\begin{equation}
    \mathcal{A}(\mathbf{x}, \theta; c)
    =
    \mathcal{A}_{\text{geo}}(\mathbf{x})
    \cdot
    \mathcal{A}_{\text{struct}}(\mathbf{x}, \theta)
    \cdot
    \mathcal{A}_{\text{scale}}(\mathbf{x}; c),
\end{equation}
with the optimal pose given by
\begin{equation}
    \mathbf{p}^*
    =
    (u^*, v^*, \theta^*)
    =
    \arg\max_{\mathbf{x}, \theta}
    \mathcal{A}(\mathbf{x}, \theta; c).
\end{equation}
Here, $\mathcal{A}_{\text{geo}}$ measures geometric clearance. Structural compatibility is represented by $\mathcal{A}_{\text{struct}}$, while $\mathcal{A}_{\text{scale}}$ measures spatial support for the target scale. Each sub-field lies in $[0,1]$. Their multiplicative combination assigns high affordance only to poses that remain compatible with both the local scene structure and the spatial requirement of the target. No learned fusion weights are required, and the complete field is constructed without training.

\paragraph{Clearance and Structural Alignment}
The clearance field $\mathcal{A}_{\text{geo}}$ retains the criterion used in the conference version. We first compute the Euclidean distance transform
\begin{equation}
    \mathcal{D}(\mathbf{x})
    =
    \inf_{\mathbf{b}\in\partial M_{valid}}
    \|\mathbf{x}-\mathbf{b}\|_2,
\end{equation}
which measures the minimum $L_2$ distance from position $\mathbf{x}$ to the boundary $\partial M_{valid}$. The normalized clearance field is
\begin{equation}
    \mathcal{A}_{\text{geo}}(\mathbf{x})
    =
    \frac{\mathcal{D}(\mathbf{x})}
    {\max_{\mathbf{x}'}\mathcal{D}(\mathbf{x}')},
\end{equation}
so that $\mathcal{A}_{\text{geo}}\in[0,1]$.

The structural field $\mathcal{A}_{\text{struct}}$ estimates orientation directly from image structure rather than from $\nabla\mathcal{D}$. We compute a multi-scale structure tensor of the background as
\begin{equation}
    \mathbf{J}(\mathbf{x})
    =
    \sum_{\sigma}
    w_\sigma\,
    G_\sigma *
    \big(
    \nabla I_{bg}(\mathbf{x})
    \nabla I_{bg}(\mathbf{x})^{\top}
    \big),
\end{equation}
where $\nabla I_{bg}\in\mathbb{R}^2$ denotes the spatial image gradient. $G_\sigma$ is a Gaussian smoothing window at scale $\sigma$ with weight $w_\sigma$, and $*$ denotes spatial convolution. The resulting tensor $\mathbf{J}(\mathbf{x})\in\mathbb{R}^{2\times2}$ is symmetric. Its eigendecomposition provides the local structural orientation
\begin{equation}
    \theta_{\text{tex}}(\mathbf{x})
    =
    \tfrac{1}{2}\,
    \mathrm{atan2}(2J_{xy},J_{xx}-J_{yy})
    +
    \tfrac{\pi}{2},
\end{equation}
and the corresponding coherence
\begin{equation}
    c_{\text{coh}}(\mathbf{x})
    =
    \frac{\lambda_1-\lambda_2}
    {\lambda_1+\lambda_2+\epsilon}.
\end{equation}
The eigenvalues satisfy $\lambda_1\geq\lambda_2\geq0$, and $\epsilon$ is a small constant for numerical stability. The orientation $\theta_{\text{tex}}$ follows the direction of least intensity variation and therefore tends to align with locally elongated structures. The coherence $c_{\text{coh}}\in[0,1]$ measures the reliability of this orientation estimate. We then define
\begin{equation}
    \mathcal{A}_{\text{struct}}(\mathbf{x}, \theta)
    =
    \max\!\big(
    0,
    \cos(2(\theta-\theta_{\text{tex}}(\mathbf{x})))
    \big)
    \cdot
    c_{\text{coh}}(\mathbf{x}),
\end{equation}
where the factor $2$ accounts for the $\pi$-periodicity of object orientation. This field provides orientation cues from elongated infrastructure such as runway and dock structures, complementing the clearance information supplied by $\mathcal{A}_{\text{geo}}$.

\paragraph{Class Scale Affordance}
The conference planner does not explicitly account for the spatial support required by different target classes. We introduce $\mathcal{A}_{\text{scale}}$ as a soft accommodation prior for this purpose. Let $(W_c,H_c)$ denote the pixel dimensions of a class-$c$ object, with reference radius
\begin{equation}
    r_c = \tfrac{1}{2}\min(W_c,H_c).
\end{equation}
The valid-region mask is eroded at several radii derived from $r_c$. The resulting masks are aggregated into
\begin{equation}
    \mathcal{A}_{\text{scale}}(\mathbf{x}; c)
    =
    \frac{1}{\sum_k\omega_k}
    \sum_k
    \omega_k
    \big(
    M_{valid}\ominus B_{s_k r_c}
    \big)(\mathbf{x}).
\end{equation}
Here, $\ominus$ denotes morphological erosion and $B_r$ is a disk structuring element with radius $r$. The factors $s_k$ determine the erosion scales, while $\omega_k$ specifies their contribution to the aggregated score. A high value indicates that the local region provides stable support for an object of the target scale. Since circular erosion is used only as a rotation-independent soft prior, the exact oriented footprint is examined separately during pose optimization. The dimensions $(W_c,H_c)$ follow the class-relative scale estimation used in the conference version.

\paragraph{Pose Optimization and Generalization}
Direct joint optimization over $(\mathbf{x},\theta)$ is unnecessary because the affordance field has a factorized structure. Both $\mathcal{A}_{\text{geo}}$ and $\mathcal{A}_{\text{scale}}$ depend only on spatial position. We therefore extract local maxima of
$\mathcal{A}_{\text{geo}}\mathcal{A}_{\text{scale}}$ as candidate locations and rank them using the complete affordance score. A small candidate budget is retained for efficient evaluation. At each candidate location, the preferred orientation is obtained in closed form as
\begin{equation}
    \theta^*
    =
    \theta_{\text{tex}}(\mathbf{x}),
\end{equation}
for which $\mathcal{A}_{\text{struct}}$ reaches $c_{\text{coh}}(\mathbf{x})$.

Each candidate is subsequently subjected to a hard geometric feasibility test. Its oriented footprint must remain sufficiently inside $M_{valid}$ without intersecting previously occupied regions. This verification complements the soft scale prior and removes geometrically invalid placements. The feasible candidate with the highest affordance score is selected as $\mathbf{p}^*$.

The conference planner is recovered when
$\mathcal{A}_{\text{struct}}\equiv1$ and
$\mathcal{A}_{\text{scale}}\equiv1$, in which case
$\mathcal{A}$ reduces to $\mathcal{A}_{\text{geo}}$. The proposed planner remains training-free and independent of the subsequent generation process. It can therefore be applied across different remote sensing modalities without retraining.

\subsection{Pose-Conditioned Decoupling ({\tt{D}})}
\label{decouple}

\paragraph{Pose-Conditioned Background Construction}

A key interface between planning and generation is how the planned pose is communicated to the generator while retaining the context of the target scene. The conference version represents the insertion region only through a bounding-box condition, which provides limited information about the spatial support determined by the planner. In \methodname{}, we instead introduce a pixel-level representation of the planned pose. Given the reference object mask, we transform it according to $\mathbf{p}^*=(u^*,v^*,\theta^*)$ and obtain the corresponding instance mask $M_{seg}$. The pose-conditioned background is then constructed as
\begin{equation}
    \tilde{I}_{bg}
    =
    \mathcal{T}(I_{bg}, M, M_{seg}),
\end{equation}
where $\mathcal{T}(\cdot)$ embeds the transformed object support within the insertion region $M$ while preserving the surrounding scene context. The resulting $\tilde{I}_{bg}$ is encoded by the VAE and introduced through the existing in-context conditioning pathway, requiring no modification to the generative architecture. This representation conveys the planned pose at pixel level and provides the generator with the environmental context needed to synthesize the object consistently with its destination scene. Compared with box-level conditioning, it therefore offers more precise control over the spatial extent of the inserted content.

The instance mask $M_{seg}$ also provides pixel-level object support for every synthesized sample. As a result, the generated images can be directly paired with their corresponding masks, extending synthetic augmentation from object detection to segmentation without additional manual annotation.

\paragraph{Observation-State Adaptation through Scene Context}
A correctly placed object must also conform to the observation characteristics of the target scene while preserving its semantic identity. The conference version addresses this issue through an explicit spectral-adaptation module based on frequency decomposition and environment-conditioned modulation. In \methodname{}, the target scene itself instead serves as evidence of the desired observation state. The pose-conditioned background $\tilde{I}_{bg}$ retains the local environmental context around the planned insertion region, allowing reference identity to interact directly with information from the destination observation during generation.

This formulation is motivated by recent evidence that pretrained diffusion models encode scene-level environmental priors useful for appearance adaptation~\cite{phongthawee2024diffusionlight,ke2024repurposing,zeng2024rgbx}. Rather than reproducing the appearance inherited from the source observation, the generator can therefore re-image the reference object according to the sensing characteristics represented by the target scene. This formulation is particularly relevant to remote sensing imagery, where appearance may vary substantially across imaging modalities and acquisition settings.

\paragraph{Decoupling Subject Identity from Observation State}
Decoupling separates the subject information that should be preserved from the scene-dependent appearance that should adapt. The reference tokens provide semantic identity through the in-context conditioning scheme introduced in Section~\ref{sec:Subject-Driven Condition Injection}. The pose-conditioned background $\tilde{I}_{bg}$ provides the spatial support together with the observation context of the destination scene. Their interaction allows the generator to retain the identity of the reference object while adapting its rendered appearance to the target observation.

Lightweight LoRA adapters specialize the generative backbone to remote sensing imagery without introducing a dedicated observation-state adaptation branch. Compared with the conference pipeline, this design replaces explicit spectral adaptation with scene-conditioned generation and integrates identity preservation directly with environment-aware appearance adaptation. As shown in Section~\ref{sec:ablation}, reintroducing the conference adaptation module on top of this formulation provides no additional benefit and can slightly reduce generation consistency.

\subsection{Manifold-Aligned Texture Assimilation ({\tt{A}})}
\label{ass}

\paragraph{Local Neighborhood Definition}
Texture compatibility is primarily determined by the local environment around the inserted object. Using distant background regions as reference may introduce statistics that are unrelated to the immediate surroundings. We therefore define a local environmental region $M_{env}$ around the insertion mask $M$. Given the planned pose $\mathbf{p}^*=(u^*,v^*,\theta^*)$ and the corresponding subject size, we obtain $M_{env}$ from the morphologically dilated mask $M_{dilated}$ as
\begin{equation}
    M_{env}=M_{dilated}\setminus M.
\end{equation}
The resulting annular region contains the neighboring background pixels immediately outside the insertion boundary and serves as the reference region for texture assimilation.

\paragraph{From Single-Statistic Matching to Manifold Alignment}
The conference version enforces local texture consistency by matching one Gram matrix from the inserted region with another from its neighborhood using an $L_2$ objective. Such a representation summarizes each region with a single second-order statistic and therefore loses variations in texture organization across spatial scales. Directly comparing the two matrices also reduces the local texture distribution to one global correspondence. We instead represent each region with a collection of spatial Gram statistics computed at multiple scales. Texture assimilation is then formulated as distribution alignment between the two collections using optimal transport. This formulation remains an inference-time guidance mechanism and introduces no additional trainable parameters.

\paragraph{Multi-Scale Spatial Gram Set}
At inference timestep $t$, let $x_t\in\mathbb{R}^{C\times h\times w}$ denote the noisy latent, where $C$ is the number of channels and $h\times w$ is its spatial resolution. For a region mask $M_r$, with $M_r$ corresponding to either $M$ or $M_{env}$, we partition the masked latent into $P_s$ spatial patches at scale $s$. A normalized Gram matrix is computed for each patch as
\begin{equation}
    G^{(s)}_p(x_t)
    =
    \frac{1}{|\Omega_p^{s}|}
    \sum_{i\in\Omega_p^{s}}
    f_i f_i^{\top}
    \in\mathbb{R}^{C\times C},
\end{equation}
where $\Omega_p^{s}$ denotes the spatial positions belonging to patch $p$ at scale $s$. The vector $f_i\in\mathbb{R}^{C}$ contains the latent features at position $i$. Normalization by $|\Omega_p^{s}|$ reduces the dependence of the Gram statistic on patch cardinality.

Aggregating the patch statistics over $S$ scales gives
\begin{equation}
\mathcal{G}(x_t;M_r)
=
\left\{
G^{(s)}_p(x_t)
\;\middle|\;
s=1,\ldots,S,\;
p=1,\ldots,P_s
\right\}.
\label{eq:gram_set}
\end{equation}

Larger spatial partitions describe broader texture organization, while finer partitions retain more localized variations. We define the resulting representations for the inserted region and its neighborhood as
\begin{equation}
    \mathcal{G}_{obj}=\mathcal{G}(x_t;M),
    \qquad
    \mathcal{G}_{env}=\mathcal{G}(x_t;M_{env}).
\end{equation}

\paragraph{Optimal-Transport Texture Loss}
We regard $\mathcal{G}_{obj}$ and $\mathcal{G}_{env}$ as empirical distributions of local second-order texture statistics. Their discrepancy is measured with an entropy-regularized optimal transport distance
\begin{equation}
    \mathcal{L}_{tex}
    =
    \mathcal{W}_{\varepsilon}
    \big(
    \mathcal{G}_{obj},
    \mathcal{G}_{env}
    \big)
    =
    \min_{\pi\in\Pi}
    \sum_{a,b}
    \pi_{ab}
    \big\|
    G_a-G_b
    \big\|_F^2
    -
    \varepsilon H(\pi),
\end{equation}
where $G_a\in\mathcal{G}_{obj}$ and $G_b\in\mathcal{G}_{env}$ denote individual Gram matrices.
The set $\Pi$ contains transport plans with uniform marginals. The pairwise transport cost is measured by the squared Frobenius distance.
We define $H(\pi)=-\sum_{a,b}\pi_{ab}\log\pi_{ab}$ as the entropy of the transport plan, with $\varepsilon>0$ controlling the
strength of entropy regularization. The transport plan $\pi$ is computed using differentiable Sinkhorn iterations. This formulation aligns the
distributions represented by the multi-scale Gram sets rather than directly matching a single regional statistic.

\paragraph{Gradient-Guided Flow Matching}
The texture objective is introduced into the sampling trajectory as energy-based guidance, following the general guidance strategy of the conference version. Let $v_\theta(x_t,t)$ denote the velocity field predicted by the flow-matching backbone~\cite{lipman2022flow} at timestep $t\in[0,1]$. We define the guided velocity as
\begin{equation}
    \hat{v}_t
    =
    v_\theta(x_t,t)
    -
    \lambda(t)
    \nabla_{x_t}
    \mathcal{L}_{tex}(x_t),
\end{equation}
where $\lambda(t)$ controls the strength of texture guidance. The guidance is activated only during the early sampling phase, when the latent representation primarily determines coarse appearance organization. It is disabled at later timesteps to preserve the subsequent synthesis of fine details.

The conference texture objective can be recovered from this formulation by setting $S=1$ and $P_1=1$, such that each region is represented by a single Gram matrix. When the entropy regularization becomes negligible, the transport objective reduces to the squared Frobenius distance between the object region and its local neighborhood
\begin{equation}
    \mathcal{L}_{tex}
    =
    \left\|
    G(x_t;M)
    -
    G(x_t;M_{env})
    \right\|_F^2.
\end{equation}

\begin{algorithm}[t]
\small
\caption{Recursive Multi-Object Synthesis}
\label{alg:batch_pipeline}
\SetAlgoNoLine
\LinesNumbered
\DontPrintSemicolon

\KwIn{
Raw background $I_{bg}^{raw}$,
subject $I_{sub}$ with mask $M_{sub}$,
target count $N$,
class $c$
}
\KwOut{Composite image $I_{final}$}

\tcp*[h]{\textbf{Stage I  Batch Layout Planning}}

$\mathcal{Q}\leftarrow[\,]$,
\quad
$M_{\mathrm{occ}}\leftarrow\mathbf{0}$\;

\For{$k\leftarrow1$ \KwTo $N$}{
    $\mathbf{p}_k
    \leftarrow
    \arg\max_{\mathbf{x},\theta}
    \mathcal{A}(\mathbf{x},\theta;c)$\;

    \tcp*[h]{subject to $M(\mathbf{x},\theta)\odot M_{\mathrm{occ}}=\mathbf{0}$}

    \If{$\mathcal{A}(\mathbf{p}_k;c)<\tau$}{
        \textbf{break}\;
    }

    $M_k\leftarrow M(\mathbf{p}_k)$\;

    $M_{\mathrm{occ}}
    \leftarrow
    M_{\mathrm{occ}}\cup M_k$\;

    $\mathrm{Append}(\mathcal{Q},\mathbf{p}_k)$\;
}

\tcp*[h]{\textbf{Stage II  Recursive Generation}}

$I_{\mathrm{curr}}
\leftarrow
I_{bg}^{raw}$\;

\While{$\mathcal{Q}\neq[\,]$}{
    $\mathbf{p}_k
    \leftarrow
    \mathrm{PopFront}(\mathcal{Q})$\;

    $M_k
    \leftarrow
    M(\mathbf{p}_k)$,
    \quad
    $M_{\mathrm{seg}}^k
    \leftarrow
    \mathrm{Transform}(M_{sub},\mathbf{p}_k)$\;

    $\tilde{I}_{\mathrm{curr}}
    \leftarrow
    \mathcal{T}(I_{\mathrm{curr}},M_k,M_{\mathrm{seg}}^k)$\;

    $I_{\mathrm{curr}}
    \leftarrow
    \mathrm{DecoupleAssimilate}
    (\tilde{I}_{\mathrm{curr}},I_{sub},\mathbf{p}_k)$\;
}

$I_{final}\leftarrow I_{\mathrm{curr}}$\;

\Return{$I_{final}$}

\end{algorithm}

\subsection{Recursive Multi-Object Synthesis}
\label{subsec:recursive_multi_object}

While \methodname{} is formulated for single-object insertion in Algorithm~\ref{alg:pda_pipeline}, practical data augmentation may require
several targets to be synthesized within the same scene. Repeatedly applying the single-object pipeline independently is not sufficient because candidate
placements may overlap and subsequent insertions cannot account for content generated in earlier steps. We therefore extend \methodname{} with the
recursive synthesis procedure summarized in Algorithm~\ref{alg:batch_pipeline}. The extension retains the same
environment-aware formulation as the single-object pipeline while coordinating the planned poses and progressively updating the scene during generation.

\paragraph{Batch Layout Planning}
Given a desired number of targets $N$, we evaluate the affordance field $\mathcal{A}(\mathbf{x},\theta;c)$ defined in Section~\ref{plan} over the
valid region. Rather than using only the highest-scoring pose, the planner iteratively constructs a pose queue
$\mathcal{Q}=\{\mathbf{p}_1,\ldots,\mathbf{p}_{N'}\}$ while maintaining a binary occupancy mask $M_{\mathrm{occ}}$. At iteration $k$, the next pose is
selected from the currently available region according to
\begin{equation}
\mathbf{p}_k
=
\arg\max_{\mathbf{x},\theta}
\mathcal{A}(\mathbf{x},\theta;c)
\quad
\mathrm{s.t.}
\quad
M(\mathbf{x},\theta)
\odot
M_{\mathrm{occ}}
=
\mathbf{0},
\label{eq:multi_pose}
\end{equation}
where $M(\mathbf{x},\theta)$ denotes the target footprint associated with the candidate pose $(\mathbf{x},\theta)$. The constraint requires the
candidate footprint to remain disjoint from the occupied region. Once $\mathbf{p}_k$ is accepted, its footprint is merged into
$M_{\mathrm{occ}}$ and excluded from subsequent planning. The procedure continues until the required number of poses has been obtained or no feasible candidate remains above the affordance threshold. The resulting poses therefore follow the same scene-compatibility criterion as the single-object planner while remaining mutually non-overlapping.

\paragraph{Recursive Generative Injection}
The planned targets are synthesized sequentially so that each insertion can respond to the current scene state. We initialize the process with
$I_{\mathrm{curr}}^{(0)}=I_{bg}^{raw}$. For a planned pose $\mathbf{p}_k$, the transformed subject mask $M_{\mathrm{seg}}^k$ defines
the object support within the insertion region $M_k$. The corresponding pose-conditioned scene is written as
\begin{equation}
\tilde{I}_{curr}^{(k)}
=
\mathcal{T}
\left(
I_{curr}^{(k-1)},
M_k,
M_{seg}^{k}
\right).
\label{eq:recursive_condition}
\end{equation}
Generation then follows the same Decoupling and Assimilation procedure as Algorithm~\ref{alg:pda_pipeline}. The reference object supplies identity
information, while the current scene provides the spatial support and observation context required for synthesis. MATA further reduces local
texture discrepancies during sampling. After each insertion, the resulting composite replaces the previous scene state and serves as the condition for the next target. The $k$-th insertion therefore depends on the environment produced by all preceding steps, allowing later targets to remain compatible with the evolving scene. This recursive formulation supports coherent multi-object synthesis and enables efficient construction of synthetic samples for downstream detection and segmentation.

\section{Experiments}

\begin{figure*}[t]
  \centering
  \includegraphics[width=0.99\textwidth]{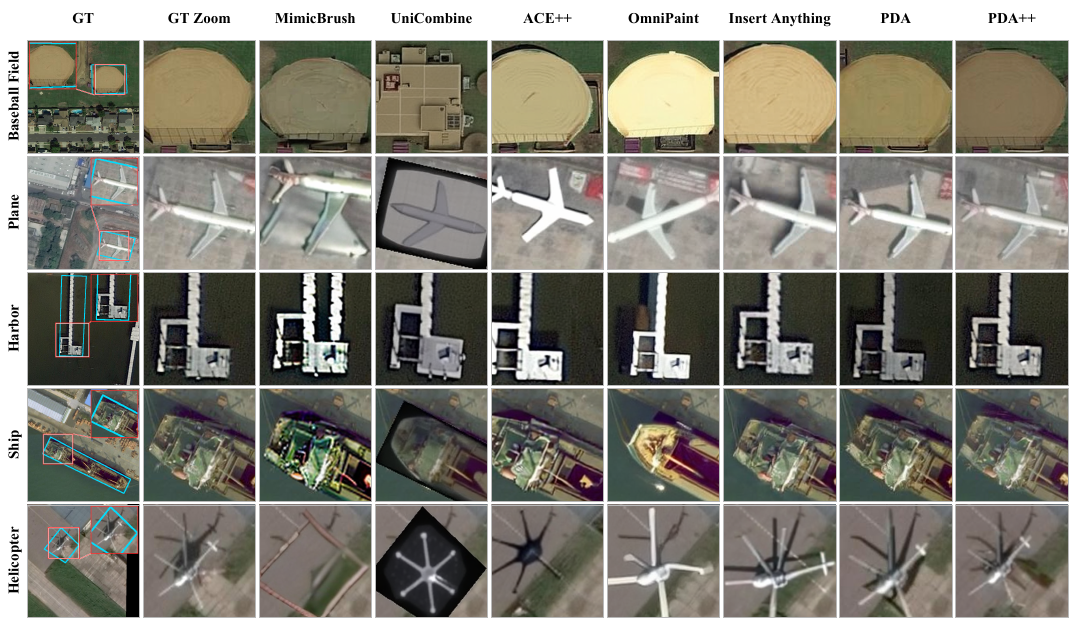}
  \caption{Qualitative comparison of object insertion. We compare our \methodname{}  with diffusion-based methods.}
  \label{fig:comparison}
\end{figure*}

\begin{table*}[t]
\centering
\caption{Quantitative comparison with state-of-the-art methods for optical remote sensing object insertion. Whole-image metrics evaluate global visual fidelity, while insertion-region metrics more directly assess the realism and environmental compatibility of the synthesized object.}
\label{tab:quantitative_comparison}
\resizebox{\textwidth}{!}{%
\setlength{\tabcolsep}{11pt}
\renewcommand{\arraystretch}{1}
\begin{tabular}{l|cccc|ccc}
\toprule
\multirow{2}{*}{Methods} & \multicolumn{4}{c|}{Whole Image} & \multicolumn{3}{c}{Insertion Region} \\
\cmidrule(lr){2-5} \cmidrule(lr){6-8}
 & PSNR $\uparrow$ & SSIM $\uparrow$ & LPIPS $\downarrow$ & FID $\downarrow$ & PSNR $\uparrow$ & SSIM $\uparrow$ & LPIPS $\downarrow$ \\
\midrule
AnyDoor~\cite{chen2024anydoor} & 24.24 & 0.8501 & 0.1040 & 21.47 & 15.70 & 0.4411 & 0.2897 \\
MimicBrush~\cite{chen2024zero} & 24.81 & \underline{0.9337} & \underline{0.0582} & 21.85 & 14.36 & 0.3676 & 0.3478 \\
Qwen Image Edit~\cite{wu2025qwen} & 18.41 & 0.7912 & 0.2044 & 46.08 & 10.72 & 0.2574 & 0.5698 \\
UniCombine~\cite{wang2025unicombine} & 24.97 & 0.8869 & 0.0781 & 22.06 & 15.89 & 0.4566 & 0.2827 \\
ACE++~\cite{mao2025ace++} & 17.44 & 0.3924 & 0.2659 & 29.24 & 15.89 & 0.4167 & 0.2757 \\
OmniPaint~\cite{yu2025omnipaint} & 24.32 & 0.8283 & 0.0871 & 18.14 & 16.83 & 0.4965 & 0.2156 \\
Insert Anything~\cite{song2025insert} & 26.12 & 0.8893 & 0.0707 & 11.54 & 18.07 & 0.5463 & 0.1561 \\
OminiControl~\cite{tan2025ominicontrol} & 25.22 & 0.8603 & 0.0839 & 12.05 & 17.90 & 0.5396 & 0.1669 \\
\midrule
PDA & \underline{28.41} & 0.8901 & 0.0601 & \underline{9.732} & \underline{20.87} & \underline{0.6249} & \underline{0.1247} \\
\methodname{} (Ours) & \textbf{32.49} & \textbf{0.9457} & \textbf{0.0252} & \textbf{6.280} & \textbf{24.91} & \textbf{0.7644} & \textbf{0.0758} \\
\bottomrule
\end{tabular}
}

\vspace{2pt}
\parbox{\linewidth}{The best results are highlighted in \textbf{bold}, and the second-best are 
\underline{underlined}.}
\end{table*}

\subsection{Optical Experiments}
\subsubsection{Experimental Setup}

We train and evaluate the optical insertion model using paired samples constructed from SAMRS~\cite{wang2023samrs} and
iSAID~\cite{waqas2019isaid}. For SAMRS, we use the FAIR1M~\cite{sun2022fair1m} subset, while iSAID is derived from
DOTA~\cite{xia2018dota}. Instance segmentation masks and oriented bounding box annotations are used to construct the training pairs. Valid target
objects are extracted and placed on a $512 \times 512$ canvas, with mild boundary smoothing applied to reduce edge aliasing. Their corresponding
regions in the original images are removed to construct the background conditions. The resulting dataset contains 19{,}163 training pairs, of which
14{,}215 are from SAMRS and 4{,}948 from iSAID. An additional 1{,}718 pairs are reserved for evaluating insertion quality.

We compare \methodname{} with representative diffusion-based editing methods, including AnyDoor~\cite{chen2024anydoor},
MimicBrush~\cite{chen2024zero}, Qwen Image Edit~\cite{wu2025qwen}, UniCombine~\cite{wang2025unicombine}, ACE++~\cite{mao2025ace++},
OmniPaint~\cite{yu2025omnipaint}, Insert Anything~\cite{song2025insert}, and OminiControl~\cite{tan2025ominicontrol}. Generation quality is evaluated at both the whole-image and insertion-region levels using PSNR, SSIM, LPIPS, and FID. As shown in the supplementary material, we further evaluate shadow and radiometric consistency to assess the physical plausibility of the synthesized targets.

\textbf{Downstream evaluation.}
We further evaluate whether the synthesized samples benefit downstrea remote sensing recognition. For oriented object detection, the real training
set is augmented with generated insertions following the conference protocol, and the change in mAP50 is measured across several oriented detectors.
We also extend the evaluation to object segmentation. The pose-conditioned construction produces an instance mask $M_{seg}$ together with each
synthesized target, allowing the generated images to be paired directly with pixel-level supervision. This enables segmentation augmentation beyond the box-level annotations supported by the conference framework. The real training set is augmented with these synthesized image and mask pairs, and
performance is evaluated on the hold-out set using mIoU and mAcc. In both settings, results are compared with training on real data alone to measure
the contribution of synthetic augmentation.

\textbf{Implementation details.}
Our framework is implemented in PyTorch and trained on two NVIDIA A100 GPU with 80GB VRAM. We use the AdamW optimizer with an initial learning rate of $1\times10^{-4}$. The model is trained for 20{,}000 steps with a batch size of 4. All images are resized to $512\times512$ during both training and inference.

\subsubsection{Optical Object Insertion Quality}\hfill\break
\hspace*{1em}\textbf{Quantitative comparison.}
As shown in Table~\ref{tab:quantitative_comparison} and Fig.~\ref{fig:comparison}, \methodname{} achieves the best performance
across all reported metrics. Compared with the conference PDA model, \methodname{} reduces whole-image FID from 9.732 to 6.28, corresponding
to a 35.5\% reduction. The improvement is more pronounced within the insertion region, where PSNR increases from 20.87 to 24.91 and LPIPS
decreases from 0.1247 to 0.0758. This difference is expected because the redesigned generation process acts primarily on the inserted content and
its local environment, whereas the unchanged background contributes substantially to whole-image measurements. The results therefore show that
the main improvement occurs within the region most directly affected by object insertion.

\textbf{Analysis of baseline behavior.}
General-purpose editors, including AnyDoor, MimicBrush, Qwen Image Edit, UniCombine, ACE++, and OmniPaint, show limited transferability from natural
image editing to overhead imagery. Remote sensing observations differ substantially in spatial organization and imaging characteristics, which are
not explicitly considered by these methods. Their generated targets may therefore remain poorly adapted to the destination scene. This effect is
particularly evident for MimicBrush. Its whole-image SSIM reaches 0.9337, whereas the insertion-region SSIM decreases to 0.3676. The large discrepancy indicates that whole-image similarity can be dominated by unchanged background content and may overestimate the quality of the synthesized target. We therefore place greater emphasis on insertion-region measurements in the following analysis.

\textbf{Discussion.}
Insertion-region results provide a more direct measure of how effectively the generated target adapts to its local environment. Among the competing
methods, Insert Anything and OminiControl achieve relatively strong local performance, with region PSNR values of 18.07 and 17.90 and SSIM values of
0.5463 and 0.5396, respectively. These models provide effective structural control, but their generation mechanisms are primarily designed for natural imagery and do not explicitly account for the observation characteristics of remote sensing data. The synthesized target can therefore preserve its reference identity without fully adapting its appearance to the destination observation. This limitation is also reflected in distributional quality. Insert Anything obtains the strongest baseline FID of 11.54, whereas \methodname{} reduces it to 6.28.

The Decoupling stage improves this adaptation by conditioning generation on the reference identity together with the pose-conditioned target background. The target background provides precise spatial support while retaining the observation context required for scene-consistent synthesis. This formulation increases region PSNR from 20.87 for PDA to 24.91 and also improves physical consistency. Assimilation further reduces residual local discrepancies by replacing the single-Gram objective of PDA with multi-scale optimal-transport alignment. Region LPIPS consequently decreases from 0.1247 to 0.0758, indicating improved local appearance compatibility. The qualitative examples in Fig.~\ref{fig:comparison} show the same tendency. Across harbor and sports-field scenes, \methodname{} produces targets that better match the destination observation and exhibit fewer visible boundary artifacts.

\begin{table*}[t]
\centering
\caption{Ablation study on the optical evaluation set. We report both whole-image and insertion-region metrics to analyze the contribution of each component to global perceptual fidelity and local insertion quality, respectively.}
\label{tab:ablation}

\resizebox{\textwidth}{!}{%
\setlength{\tabcolsep}{13pt}
\renewcommand{\arraystretch}{1}
\begin{tabular}{lccccccc}
\toprule
\multirow{2}{*}{Method Variant}
& \multicolumn{4}{c}{Whole Image}
& \multicolumn{3}{c}{Insertion Region} \\
\cmidrule(lr){2-5} \cmidrule(lr){6-8}
& PSNR $\uparrow$
& SSIM $\uparrow$
& LPIPS $\downarrow$
& FID $\downarrow$
& PSNR $\uparrow$
& SSIM $\uparrow$
& LPIPS $\downarrow$ \\
\midrule

PDA
& 28.41 & 0.8901 & 0.0601 & 9.732
& 20.87 & 0.6249 & 0.1247 \\

PDA (FLUX.2 backbone)
& 29.56 & 0.9278 & 0.0332 & 7.159
& 21.06 & 0.6102 & 0.1034 \\

\midrule

FLUX.2
& 29.44 & 0.9288 & 0.0306 & 6.992
& 20.72 & 0.5943 & 0.1078 \\

+ ASA
& 29.52 & 0.9276 & 0.0333 & 7.189
& 21.00 & 0.6077 & 0.1045 \\

+ NATA
& 29.52 & 0.9289 & 0.0304 & 6.946
& 20.83 & 0.5969 & 0.1062 \\

+ MATA
& 29.51 & 0.9289 & 0.0304 & 6.925
& 20.79 & 0.5945 & 0.1076 \\

\midrule

FLUX.2 (seg.)
& \underline{32.46}
& \underline{0.9456}
& \underline{0.0253}
& \underline{6.403}
& \underline{24.88}
& \underline{0.7550}
& \underline{0.0763} \\

+ ASA
& 31.93 & 0.9407 & 0.0261 & 6.541
& 24.76 & 0.7489 & 0.0771 \\

+ NATA
& \underline{32.46}
& \underline{0.9456}
& \underline{0.0253}
& 6.404
& 24.83 & 0.7541 & 0.0770 \\

+ MATA (PDA++ Ours)
& \textbf{32.49}
& \textbf{0.9457}
& \textbf{0.0252}
& \textbf{6.280}
& \textbf{24.91}
& \textbf{0.7644}
& \textbf{0.0758} \\

\bottomrule
\end{tabular}
}

\vspace{-3pt}
\end{table*}

\subsubsection{Ablation Study on Optical Insertion}
\label{sec:ablation}\hfill\break
\hspace*{1em}\textbf{Component analysis.}
We evaluate the major design choices on the optical test set (Table~\ref{tab:ablation}) and separately examine the effect of the backbone
upgrade. In the table, seg. denotes the pose-conditioned background construction, where $M_{seg}$ represents the planned object support at pixel
level. Replacing only the conference backbone with FLUX.2 reduces whole-image FID from 9.732 to 7.159 and LPIPS from 0.0601 to 0.0332.
The improvement within the insertion region is much smaller. Region PSNR changes from 20.87 to 21.06, while SSIM decreases from 0.6249 to 0.6102.
This suggests that the stronger backbone mainly improves global image quality but does not by itself provide substantially better control over the inserted content. The final gains of \methodname{} therefore cannot be explained by the backbone replacement alone.

A much larger improvement is obtained after introducing the pose-conditioned background. Compared with plain FLUX.2, region PSNR increases from 20.72 to 24.88, SSIM rises from 0.5943 to 0.7550, and LPIPS decreases from 0.1078 to 0.0763. Whole-image FID also improves from 6.992 to 6.403. These changes show that the pose-conditioned representation affects not only the spatial extent of the synthesized target but also the quality of its local integration. By encoding the planned object support while retaining the surrounding scene context, it provides a more informative condition for adapting the target to the destination observation. The dominant improvement in local insertion quality therefore comes from the redesigned conditioning rather than from the backbone upgrade.

Texture assimilation provides a further refinement once reliable spatial conditioning has been established. Applying MATA directly to plain FLUX.2
only slightly changes the results, reducing FID from 6.992 to 6.925 with little improvement in the insertion region. Under pose-conditioned input,
however, MATA further reduces FID from 6.403 to 6.280, increases region SSIM from 0.7550 to 0.7644, and lowers LPIPS from 0.0763 to 0.0758. In comparison, NATA remains close to the configuration without texture guidance, with an FID of 6.404 and a region PSNR of 24.83. This suggests that the multi-scale distribution alignment introduced by MATA is better suited to refining the remaining appearance discrepancy after the object support has been established. Its contribution is smaller than that of the pose-conditioned background, but it consistently improves the final configuration and yields the best overall results in the table.

\textbf{Effect of the conference appearance adaptation module.}
We further reintroduce ASA to examine whether explicit appearance adaptation remains necessary with FLUX.2. Without pose-conditioned input, ASA slightly improves region PSNR from 20.72 to 21.00 and SSIM from 0.5943 to 0.6077, while whole-image FID increases from 6.992 to 7.189. This indicates that the module can modify the local appearance to some extent, but the resulting change does not translate into a consistent improvement in overall generation quality.

The same tendency becomes more evident after pose-conditioned input is introduced. Adding ASA increases FID from 6.403 to 6.541, while region PSNR
decreases from 24.88 to 24.76 and SSIM from 0.7550 to 0.7489. The pose-conditioned background already exposes the generator to the planned
object support together with the observation context of the destination scene. Additional explicit appearance adaptation therefore provides limited
benefit under the redesigned formulation and can interfere with the scene-conditioned generation process. We consequently omit ASA from the
final \methodname{} configuration.

Overall, the ablation results clarify the contribution of the main design choices. The backbone upgrade primarily improves whole-image fidelity, while the pose-conditioned construction accounts for the dominant improvement within the insertion region. MATA then provides an additional refinement by reducing the residual discrepancy between the synthesized object and its local surroundings. The full \methodname{} therefore improves both global generation quality and local insertion fidelity without relying on the conference appearance adaptation module.

\textbf{Physical consistency.}
Additional evaluations of shadow and radiometric consistency are provided in the supplementary material and show the same trend as the visual quality results. The pose-conditioned background substantially improves physical consistency, while reintroducing ASA provides no further benefit under the proposed configuration.

\begin{figure}[t]
\centering
\includegraphics[width=0.48\textwidth]{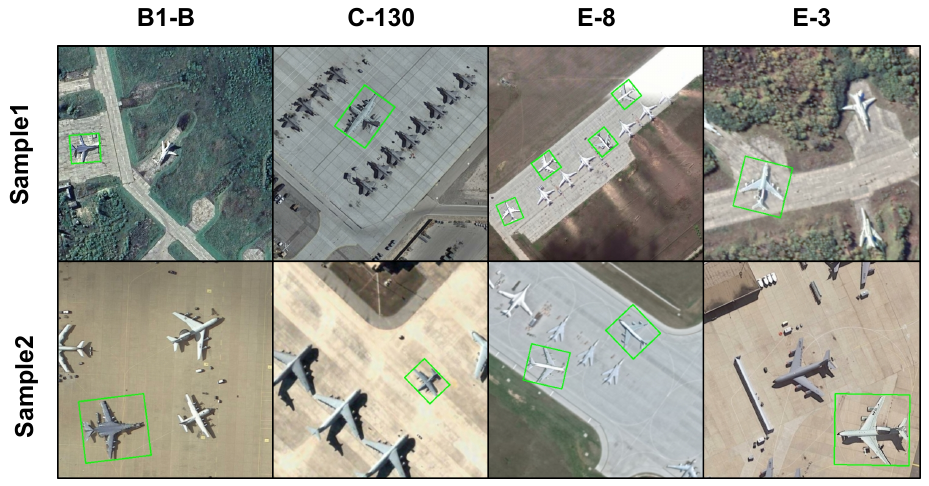}
\caption{Case studies of synthetic data under different scenarios.}
\label{fig:case_study}
\vspace{-12pt}
\end{figure}
\subsubsection{Downstream Optical Object Recognition}

\begin{table*}[t]
\centering
\scriptsize
\caption{Downstream oriented object detection performance (mAP50 in \%) on the MAR20-11-FewShot benchmark under zero-shot optical synthesis. CP denotes Copy-Paste, CM denotes CutMix, and OC denotes OminiControl. $\Delta$ Over Real denotes the improvement of \methodname{} over the real-data baseline.}
\label{tab:detection_results}
\resizebox{\textwidth}{!}{%
\setlength{\tabcolsep}{12pt}
\begin{tabular}{lccccccc}
\toprule
Detector & Real & +CP & +CM & +OC & PDA & \methodname{} (Ours) & $\Delta$ Over Real \\
\midrule
Rotated R-CNN~\cite{yang2020rotated}  
& 58.13 & 73.14 & 68.24 & 55.22 & \underline{77.88} & \textbf{78.24} & +20.11 \\

Oriented R-CNN~\cite{xie2021oriented} 
& 68.51 & 78.01 & 71.07 & 70.04 & \underline{78.59} & \textbf{78.71} & +10.20 \\

S2ANet~\cite{han2021align}            
& 55.67 & 71.86 & 62.77 & 62.36 & \underline{77.36} & \textbf{78.13} & +22.46 \\

YOLO26~\cite{sapkota2025yolo26}       
& 63.51 & 75.54 & 76.79 & 70.01 & \underline{80.26} & \textbf{81.52} & +18.01 \\

Avg. 
& 61.46 & 74.64 & 69.72 & 64.41 & \underline{78.52} & \textbf{79.15} & +17.69 \\
\bottomrule
\end{tabular}
}

\vspace{2pt}
\parbox{\linewidth}{The best results are highlighted in \textbf{bold}, and the second-best are \underline{underlined}.}
\end{table*}

\begin{table*}[t]
\centering
\scriptsize
\caption{Comparison of synthesis strategies across segmentation backbones (\%). Best in bold, second best underlined, compared among strategies within each column.}
\label{tab:segmentation_results}
\resizebox{\textwidth}{!}{%
\setlength{\tabcolsep}{12pt}
\begin{tabular}{lcccccccc}
\toprule
\multirow{2}{*}{Strategy}
& \multicolumn{2}{c}{BiSeNetV2~\cite{yu2021bisenet}}
& \multicolumn{2}{c}{PIDNet~\cite{xu2023pidnet}}
& \multicolumn{2}{c}{SegNeXt~\cite{guo2022segnext}}
& \multicolumn{2}{c}{SegFormer~\cite{xie2021segformer}} \\
\cmidrule(lr){2-3}
\cmidrule(lr){4-5}
\cmidrule(lr){6-7}
\cmidrule(lr){8-9}
& mIoU & mAcc
& mIoU & mAcc
& mIoU & mAcc
& mIoU & mAcc \\
\midrule
Baseline 
& 39.26 & 50.71
& 28.49 & 39.36
& 62.39 & 74.02
& 60.26 & 72.11 \\

+ CP
& \underline{53.98} & \underline{65.82}
& \underline{55.13} & \underline{66.55}
& \underline{70.10} & \underline{79.89}
& \underline{69.40} & \underline{80.00} \\

+ CM
& 48.89 & 61.06
& 47.09 & 59.18
& 65.10 & 75.82
& 68.33 & 79.00 \\

\methodname{} (Ours)
& \textbf{65.11} & \textbf{74.45}
& \textbf{65.18} & \textbf{75.49}
& \textbf{70.84} & \textbf{80.57}
& \textbf{71.58} & \textbf{81.01} \\
\bottomrule
\end{tabular}
}

\vspace{2pt}
\parbox{\linewidth}{The best results are highlighted in \textbf{bold}, and the second-best are \underline{underlined}.}
\end{table*}

To evaluate the practical value of the synthesized samples, we conduct few-shot oriented object recognition on MAR20-11-FewShot, a benchmark derived
from MAR20~\cite{wenqi2024mar20}. The benchmark contains 11 strategic airframe categories with 30 real images per category, resulting in 330
training images and a highly data-limited setting. The generator is trained only on pairs constructed from SAMRS/FAIR1M and iSAID/DOTA and is applied directly to MAR20 backgrounds. The resulting synthesis is therefore performed in a \emph{zero-shot} setting with respect to MAR20. We construct the augmented training set using the recursive multi-object synthesis procedure described in Section~\ref{subsec:recursive_multi_object}. Copy-Paste, CutMix~\cite{burgert2025label}, and OminiControl are included as augmentation baselines. Performance is evaluated with Rotated R-CNN~\cite{yang2020rotated}, Oriented R-CNN~\cite{xie2021oriented}, S2ANet~\cite{han2021align}, and YOLO26~\cite{sapkota2025yolo26}.

\begin{figure*}[t]
\centering
\includegraphics[width=0.98\textwidth]{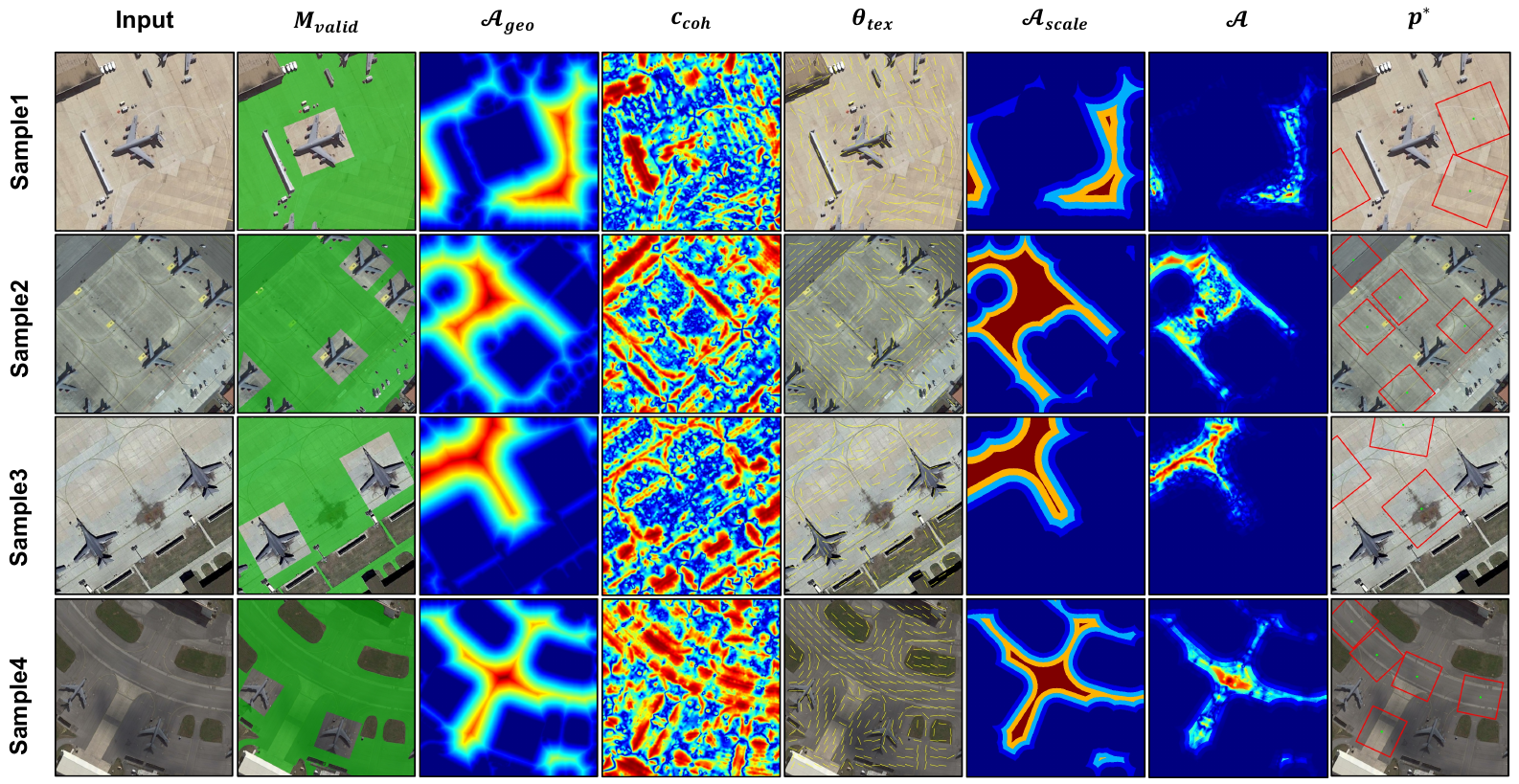}
\caption{Representative examples of affordance-aware layout planning on MAR20 backgrounds. From left to right, we show the input background, semantic valid region $M_{valid}$, clearance field $\mathcal{A}_{geo}$, structural coherence $c_{coh}$, estimated structural orientation $\theta_{tex}$, class-scale affordance $\mathcal{A}_{scale}$, fused affordance field $\mathcal{A}$, and the resulting planned poses. The planner concentrates candidate placements on spatially feasible and structurally compatible regions before recursive object generation.}
\label{fig:plan_examples}
\end{figure*}

Fig.~\ref{fig:plan_examples} shows representative planning results used to construct the synthetic training set. Semantic parsing first restricts the candidate locations in each MAR20 background to the valid region $M_{valid}$. The affordance field $\mathcal{A}$ then integrates geometric
clearance with scene structure and target-scale information to estimate suitable insertion locations. Regions with high affordance scores provide
scene-compatible support for the target and are selected as candidate poses. These poses are subsequently passed to the recursive generation stage, which inserts multiple targets into each background while preventing spatial overlap between accepted placements.

\begin{figure}[t]
\centering
\includegraphics[width=0.98\columnwidth]{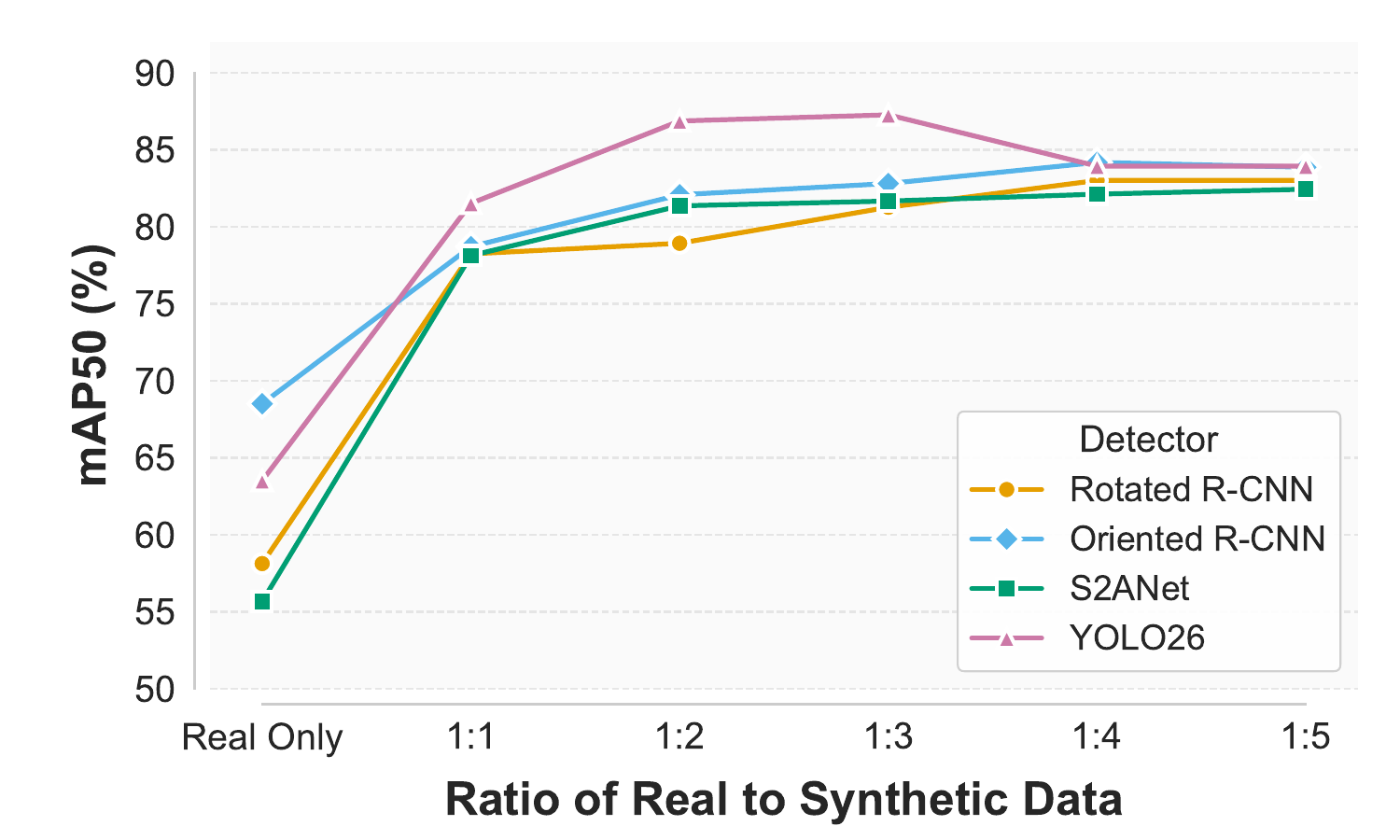}
\caption{Effect of synthetic data ratio on optical recognition performance.}
\vspace{-6pt}
\label{fig:synthetic_ratio}
\end{figure}

\begin{figure}[t]
\centering
\includegraphics[width=0.98\columnwidth]{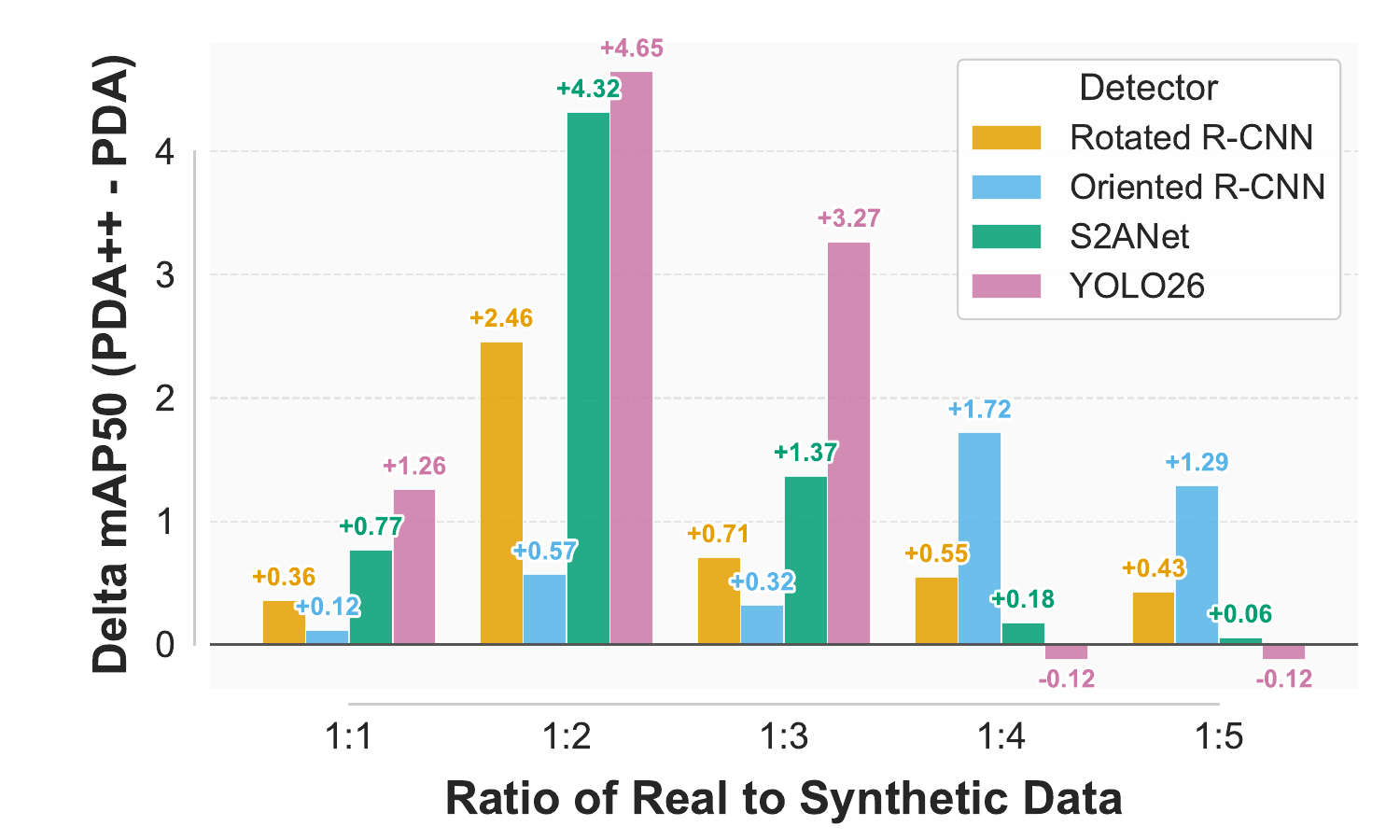}
\caption{mAP50 improvement of \methodname{} over PDA under different synthetic data ratios on optical object recognition.}
\vspace{-6pt}
\label{fig:synthetic_ratio2}
\end{figure}

\textbf{Results.}
As shown in Table~\ref{tab:detection_results}, Copy-Paste and CutMix provide only modest improvements because the inserted targets are not adapted to the
surrounding environment. OminiControl shows less stable downstream performance and underperforms the real-data baseline on some detectors. For
example, its mAP50 on Rotated R-CNN is 55.22, compared with 58.13 for real-data training. Although OminiControl uses a similar generative backbone,
this result suggests that visual plausibility alone does not guarantee effective augmentation. Synthetic targets that remain incompatible with the
destination scene may introduce unreliable supervision. In contrast, \methodname{} improves all four detectors and raises the average mAP50
from 61.46 to 79.15, corresponding to an absolute gain of 17.69 points and a relative improvement of 28.8\%. It also consistently outperforms the
conference PDA framework. Similar improvements across different detector architectures indicate that the benefit mainly originates from the synthetic
training samples rather than from a particular detector.

\textbf{Effect of synthetic ratio.}
We vary the synthetic-to-real ratio for PDA and \methodname{} in Fig.~\ref{fig:synthetic_ratio} and~\ref{fig:synthetic_ratio2}.
Introducing synthetic data substantially improves all detectors over the real-data baseline, while \methodname{} generally provides higher accuracy
than PDA across the evaluated ratios. The advantage is particularly clear when only a limited amount of synthetic data is used. At a ratio of 1:2,
\methodname{} exceeds PDA by 4.65 mAP50 on YOLO26 and by 4.32 points on S2ANet, indicating that the generated samples provide effective
supervision even at relatively low synthetic ratios.

For YOLO26, performance has already entered its optimum range at a ratio of 1:2, after which additional synthetic samples provide little further benefit. The small negative differences of 0.12 points relative to PDA at ratios of 1:4 and 1:5 therefore occur after performance has saturated and do not indicate a meaningful loss in absolute detection accuracy. A similar saturation tendency is observed for the other detectors as the amount of
synthetic data increases. Overall, the results suggest that \methodname{} achieves most of its downstream benefit with a moderate amount of synthetic data, while further increasing the synthetic proportion leads to diminishing returns.

\subsubsection{Downstream Optical Instance Segmentation}

The pose-conditioned construction associates each synthesized target with a pixel-level mask $M_{seg}$ that directly represents its spatial support. The generated samples can therefore provide segmentation supervision without additional manual annotation, extending the box-level augmentation supported by the conference framework. We evaluate this capability on the same MAR20-11-FewShot images. Real segmentation annotations are obtained from the FineGrip panoptic annotations of MAR20~\cite{finegrip}. The synthetic images reuse the zero-shot insertions described above and are paired with their corresponding $M_{seg}$ masks.

For segmentation, each model is trained using real and synthesized samples at a fixed real-to-synthetic ratio of 1:3. Evaluation is performed on the
held-out real test set using mIoU and mAcc. We consider BiSeNetV2~\cite{yu2021bisenet}, PIDNet~\cite{xu2023pidnet}, SegNeXt~\cite{guo2022segnext}, and SegFormer~\cite{xie2021segformer}. As shown in Table~\ref{tab:segmentation_results}, \methodname{} improves performance for every segmentation architecture and consistently surpasses Copy-Paste and CutMix. The improvement is especially pronounced for BiSeNetV2, where mIoU increases from 39.26 to 65.11. PIDNet shows a similar increase from 28.49 to 65.18. SegNeXt and SegFormer also benefit consistently from the synthesized training samples. These results confirm that the automatically associated object masks provide effective pixel-level supervision and extend the utility of \methodname{} to segmentation augmentation.

\begin{table*}[t]
\centering
\caption{Quantitative comparison on HRSID for SAR object insertion. We report both whole-image and insertion-region metrics to evaluate global generation fidelity and local insertion quality, respectively.}
\label{tab:sar_generation}
\resizebox{\textwidth}{!}{%
\setlength{\tabcolsep}{13pt}
\renewcommand{\arraystretch}{1}
\begin{tabular}{l|cccc|ccc}
\toprule
\multirow{2}{*}{Method Variant} & \multicolumn{4}{c|}{Whole Image} & \multicolumn{3}{c}{Insertion Region} \\
\cmidrule(lr){2-5} \cmidrule(lr){6-8}
 & PSNR $\uparrow$ & SSIM $\uparrow$ & LPIPS $\downarrow$ & FID $\downarrow$ & PSNR $\uparrow$ & SSIM $\uparrow$ & LPIPS $\downarrow$ \\
\midrule
PDA 
& 25.16 & 0.7643 & 0.0682 & 20.03 
& 15.36 & 0.6942 & 0.0477 \\

Flux.2 (seg.) 
& \underline{27.19} & \textbf{0.8666} & \underline{0.0437} & 18.76 
& \underline{16.17} & \underline{0.7280} & 0.0433 \\

\quad + NATA 
& 27.17 & \underline{0.8660} & 0.0446 & \underline{18.70} 
& 16.14 & 0.7274 & \underline{0.0428} \\

\methodname{} (Ours) 
& \textbf{27.26} & \textbf{0.8666} & \textbf{0.0431} & \textbf{18.46} 
& \textbf{16.31} & \textbf{0.7343} & \textbf{0.0408} \\
\bottomrule
\end{tabular}
}

\vspace{2pt}
\parbox{\linewidth}{The best results are highlighted in \textbf{bold}, and the second-best are \underline{underlined}.}
\end{table*}

\subsection{SAR Experiments}
\subsubsection{Experimental Setup}

\begin{table*}[t]
\centering
\scriptsize
\caption{Downstream ship detection performance (mAP50 in \%) on SSDD under the zero-shot SAR augmentation setting. The generator is trained on HRSID and directly applied to SSDD without additional fine-tuning. CP denotes Copy-Paste, CM denotes CutMix, and OC denotes OminiControl. $\Delta$ Over Real denotes the improvement of PDA++ over the real-data baseline.}
\label{tab:sar_detection}
\resizebox{\textwidth}{!}{%
\setlength{\tabcolsep}{12pt}
\renewcommand{\arraystretch}{1}
\begin{tabular}{lccccccc}
\toprule
Detector & Real & +CP & +CM & +OC & PDA & \methodname{} (Ours) & $\Delta$ Over Real \\
\midrule
Rotated R-CNN~\cite{yang2020rotated}  
& 77.91 & 77.01 & 66.01 & 78.84 & \underline{78.86} & \textbf{79.13} & +1.22 \\

Oriented R-CNN~\cite{xie2021oriented}  
& 80.47 & 80.00 & 80.00 & 87.97 & \underline{88.60} & \textbf{88.92} & +8.45 \\

S2ANet~\cite{han2021align}         
& 77.36 & 65.44 & 75.29 & 77.86 & \underline{78.42} & \textbf{79.05} & +1.69 \\

YOLO26~\cite{sapkota2025yolo26}         
& 79.62 & 78.93 & 79.30 & 81.57 & \underline{84.28} & \textbf{84.67} & +5.05 \\

Avg.           
& 78.84 & 75.35 & 75.15 & 81.56 & \underline{82.54} & \textbf{82.94} & +4.10 \\
\bottomrule
\end{tabular}
}

\vspace{2pt}
\parbox{\linewidth}{The best results are highlighted in \textbf{bold}, and the second-best are \underline{underlined}.}
\end{table*}

\begin{table*}[t]
\centering
\scriptsize
\caption{Comparison of synthesis strategies across segmentation backbones for SAR object segmentation (\%). Best in bold, second best underlined, compared among strategies within each column.}
\label{tab:sar_segmentation_results}
\resizebox{\textwidth}{!}{%
\setlength{\tabcolsep}{12pt}
\renewcommand{\arraystretch}{1}
\begin{tabular}{lcccccccc}
\toprule
\multirow{2}{*}{Strategy}
& \multicolumn{2}{c}{BiSeNetV2~\cite{yu2021bisenet}}
& \multicolumn{2}{c}{PIDNet~\cite{xu2023pidnet}}
& \multicolumn{2}{c}{SegNeXt~\cite{guo2022segnext}}
& \multicolumn{2}{c}{SegFormer~\cite{xie2021segformer}} \\
\cmidrule(lr){2-3}
\cmidrule(lr){4-5}
\cmidrule(lr){6-7}
\cmidrule(lr){8-9}
& mIoU & mAcc
& mIoU & mAcc
& mIoU & mAcc
& mIoU & mAcc \\
\midrule
Baseline
& \underline{67.85} & 78.39
& 62.58 & 74.30
& 69.65 & 78.25
& 70.66 & 78.32 \\

+ CP
& 67.63 & 81.89
& \underline{64.13} & \underline{80.11}
& \underline{71.89} & \underline{80.21}
& 71.77 & 80.77 \\

+ CM
& \underline{67.85} & \underline{82.11}
& 64.05 & 80.09
& 70.36 & 79.92
& \underline{71.83} & \underline{81.04} \\

\methodname{} (Ours)
& \textbf{72.26} & \textbf{84.63}
& \textbf{69.82} & \textbf{83.04}
& \textbf{73.79} & \textbf{82.38}
& \textbf{74.28} & \textbf{81.95} \\
\bottomrule
\end{tabular}
}

\vspace{2pt}
\parbox{\linewidth}{The best results are highlighted in \textbf{bold}, and the second-best are \underline{underlined}.}
\end{table*}

We further evaluate \methodname{} on SAR imagery, whose image formation and local statistics differ substantially from optical observations. HRSID ~\cite{wei2020hrsid} is used as the main benchmark for ship insertion. The images are divided into $256\times256$ patches, producing 3{,}600 training samples and 500 evaluation samples. For downstream evaluation, the model trained on HRSID is directly applied to SSDD~\cite{zhang2021sar} without additional training on the target dataset. SSDD contains 50 training images, 100 validation images, and 250 test images, with spatial resolutions ranging from 1 to 15~m. We additionally use WHU-OPT-SAR~\cite{li2022mcanet} to evaluate insertion on forest regions with irregular boundaries. Forest regions are removed according to their segmentation annotations and reconstructed from reference observations. This setting contains 8{,}616 training patches and 500 test patches.

Unless otherwise specified, the SAR experiments follow the implementation protocol used for optical imagery. Insertion quality is measured using PSNR, SSIM, LPIPS, and FID at the whole-image and insertion-region levels. Detection performance is reported with mAP50, while segmentation is evaluated using mIoU and mAcc. Each synthesized sample is paired with the pose-conditioned mask $M_{seg}$ and can therefore provide pixel-level supervision without additional annotation. For segmentation, the models are trained with a fixed real-to-synthetic ratio of 1:3 and evaluated on held-out real images.

\subsubsection{SAR Object Insertion Quality}

\begin{figure}[t]
\centering
\includegraphics[width=\columnwidth]{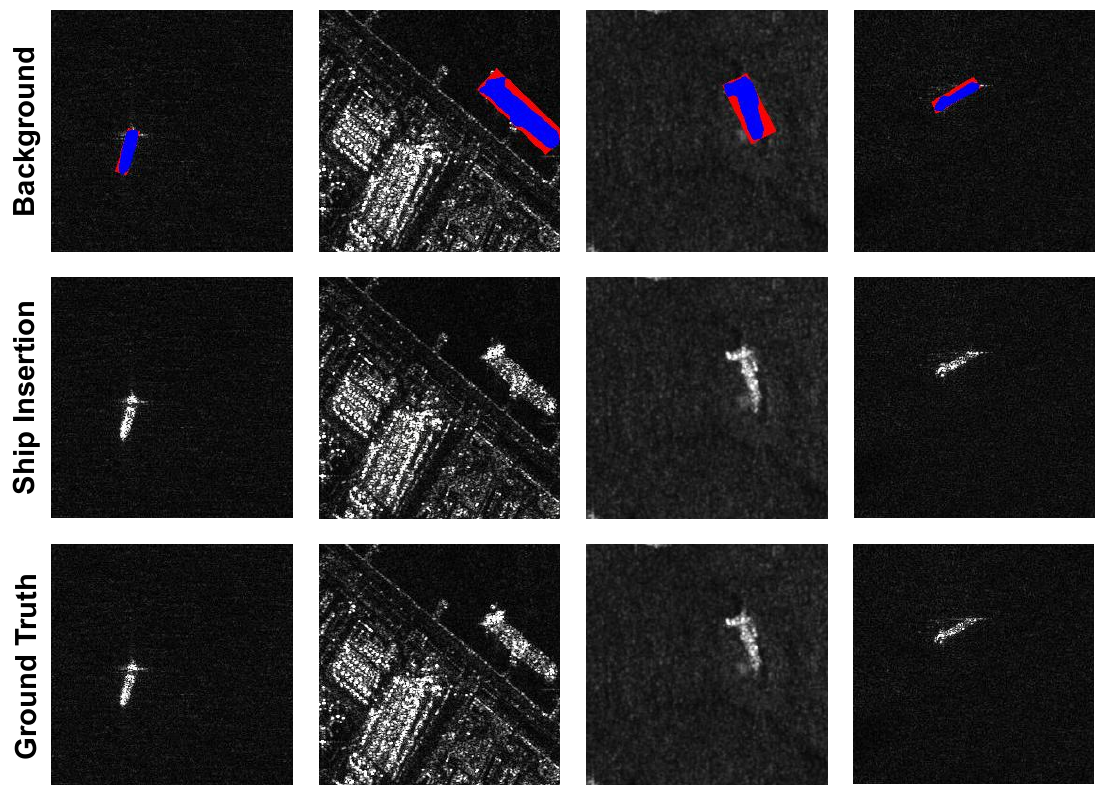}
\caption{Representative examples of SAR ship insertion on HRSID. Placement
masks are shown above the generated results, with the corresponding
ground-truth images provided for reference. \methodname{} synthesizes ships
at the specified locations while maintaining compatibility with the local
SAR appearance.}
\vspace{-6pt}
\label{fig:sar_ship_examples}
\end{figure}

We first examine insertion quality on HRSID. Representative examples are shown in Fig.~\ref{fig:sar_ship_examples}, and quantitative results are
reported in Table~\ref{tab:sar_generation}. This experiment evaluates whether the proposed formulation remains effective when the target observation
follows a substantially different imaging process from optical imagery.

\methodname{} achieves the strongest overall performance on HRSID. The whole-image results reach 27.26 PSNR, 0.8666 SSIM, 0.0431 LPIPS, and
18.46 FID. Within the insertion region, PSNR reaches 16.31, SSIM reaches 0.7343, and LPIPS decreases to 0.0408. Compared with the conference PDA model, whole-image FID decreases from 20.03 to 18.46, while region PSNR increases from 15.36 to 16.31. The improvement indicates that the redesigned generation process remains effective under the observation characteristics of SAR imagery.

The ablation results follow a trend similar to that observed in the optical experiments. Introducing the pose-conditioned background increases region PSNR from 15.36 to 16.17 and SSIM from 0.6942 to 0.7280 relative to PDA. The pixel-level support provides more precise spatial control, while the surrounding SAR scene supplies the observation context needed during generation. MATA provides a further improvement over this configuration. The conference NATA objective slightly reduces whole-image FID but changes region PSNR from 16.17 to 16.14. In contrast, MATA achieves an FID of 18.46 together with a region PSNR of 16.31. This result suggests that distribution alignment based on multi-scale texture statistics is more effective for reducing residual local discrepancies than the original single-Gram objective.

These results show that the redesigned Decoupling and Assimilation stages also remain effective for SAR imagery. The pose-conditioned representation provides the scene information required for target adaptation, while MATA further improves compatibility with the local background. No modality-specific redesign of the insertion pipeline is required.

\subsubsection{SAR Downstream Detection}

To evaluate the usefulness of the synthesized SAR samples for downstream recognition, we conduct few-shot ship detection on SSDD. The generator is trained on HRSID and directly applied to SSDD, giving a zero-shot cross-dataset evaluation. The resolution variation in SSDD also provides a test of generalization across spatial scales. We compare training on real data alone with Copy-Paste, CutMix, OminiControl, and \methodname{} using Rotated R-CNN, Oriented R-CNN, S2ANet, and YOLO26.

As shown in Table~\ref{tab:sar_detection}, \methodname{} achieves the best performance for every detector. The average mAP50 increases from 78.84 with real-data training to 82.94, corresponding to an improvement of 4.10 points. \methodname{} also outperforms the conference PDA model for all four detectors. Copy-Paste and CutMix provide less stable gains and can decrease accuracy relative to real-data training, indicating that direct composition does not reliably preserve compatibility between the synthesized ship and the SAR background. OminiControl performs more competitively but remains below \methodname{} across the evaluated detectors. Since no SSDD images are used to train the generator, the improvement also shows that the synthesis model transfers across datasets without dataset-specific adaptation. Its performance under the resolution variation of SSDD further supports generalization across spatial scales.

\subsubsection{SAR Downstream Segmentation}

We further evaluate ship segmentation on SSDD using the same zero-shot synthetic samples. Each generated image is paired with its automatically obtained mask $M_{seg}$, while the real SSDD images use their native pixel-level annotations. Following the optical protocol, each segmentation model is trained at a fixed real-to-synthetic ratio of 1:3 and evaluated on the real test set. We use BiSeNetV2, PIDNet, SegNeXt, and SegFormer for this evaluation.

As shown in Table~\ref{tab:sar_segmentation_results}, \methodname{} improves all four segmentation models and consistently surpasses Copy-Paste and CutMix. On BiSeNetV2, mIoU increases from 67.85 to 72.26. PIDNet improves from 62.58 to 69.82, while SegNeXt increases from 69.65 to 73.79. SegFormer shows a similar improvement from 70.66 to 74.28. The corresponding mAcc results exhibit the same overall tendency. In comparison, Copy-Paste and CutMix provide limited gains and can reduce segmentation accuracy in this setting. The results indicate that the masks associated with the synthesized targets provide effective pixel-level supervision in SAR imagery as well as in the optical setting.

\textbf{Beyond rigid targets.}
As detailed in the supplementary material, we further evaluate forest insertion on WHU-OPT-SAR~\cite{li2022mcanet} to examine whether the proposed framework generalizes to targets with irregular boundaries. The dataset has a spatial resolution of 5~m, providing a setting that differs substantially from the ship experiments. The pose-conditioned formulation reduces whole-image FID from 30.88 for PDA to 15.55, while MATA provides a further improvement in overall fidelity. These results indicate that the proposed formulation is not restricted to rigid targets and remains effective under notable changes in target geometry and image resolution.

\subsection{Limitations}
The current framework assumes access to pixel-level object masks for constructing the pose-conditioned background and generating segmentation supervision. Although such masks can be obtained from existing segmentation models, this requirement may limit applicability to datasets that provide only bounding-box annotations or weak labels. Developing mask-free conditioning or automatically estimating object support from boxes, points, or reference images would therefore make \methodname{} more broadly applicable. In addition, the current framework is validated on RGB optical and SAR imagery. Extending it to multispectral or hyperspectral observations requires latent representations that preserve inter-band correlations and material-dependent spectral signatures, which are not fully supported by standard RGB-oriented encoders. Addressing these limitations will further improve the flexibility of \methodname{} across annotation settings and sensing modalities.

\section{Conclusion}

We presented \methodname{}, a unified framework for remote sensing object insertion that progressively adapts synthesized targets to their destination scenes. The proposed formulation improves scene-aware placement through the Affordance Field and uses pose-conditioned scene context to guide target generation under the desired observation. MATA further reduces residual local appearance discrepancies through multi-scale texture alignment. The same pose-conditioned construction also provides pixel-level masks for segmentation augmentation, while recursive synthesis extends the framework to multi-object scene generation.

Extensive experiments on optical and SAR imagery validate both insertion quality and downstream utility. On the optical benchmark, \methodname{}
achieves a whole-image FID of 6.28 and increases the average few-shot detection mAP$_{50}$ from 61.46 to 79.15, corresponding to an improvement
of 17.69 points and a relative gain of 28.8\%. On SAR imagery, the proposed method improves SSDD ship detection by 4.10 mAP50 points over real-data training, while consistent gains are observed in segmentation and cross-dataset evaluation. These results show that adapting synthesized targets to the destination environment provides an effective basis for remote sensing generation and synthetic data augmentation.

\bibliographystyle{IEEEtran}
\bibliography{refs}

@inproceedings{
goktepe2025ecomapper,
title={EcoMapper: Generative Modeling for Climate-Aware Satellite Imagery},
author={Muhammed Goktepe and Amir hossein Shamseddin and Erencan Uysal and Javier Muinelo Monteagudo and Lukas Drees and Aysim Toker and Senthold Asseng and Malte von Bloh},
booktitle={Forty-second International Conference on Machine Learning},
year={2025},
url={https://openreview.net/forum?id=YUtJsxQjv3}
}

@article{mei2024comprehensive,
  title={A comprehensive study on the robustness of deep learning-based image classification and object detection in remote sensing: Surveying and benchmarking},
  author={Mei, Shaohui and Lian, Jiawei and Wang, Xiaofei and Su, Yuru and Ma, Mingyang and Chau, Lap-Pui},
  journal={Journal of Remote Sensing},
  volume={4},
  pages={0219},
  year={2024},
  publisher={AAAS}
}

@inproceedings{khanna2024diffusionsat,
 author = {Khanna, Samar and Liu, Patrick and Zhou, Linqi and Meng, Chenlin and Rombach, Robin and Burke, Marshall and Lobell, David and Ermon, Stefano},
 booktitle = {International Conference on Learning Representations},
 editor = {B. Kim and Y. Yue and S. Chaudhuri and K. Fragkiadaki and M. Khan and Y. Sun},
 pages = {5586--5604},
 title = {DiffusionSat: A Generative Foundation Model for Satellite Imagery},
 url = {https://proceedings.iclr.cc/paper_files/paper/2024/file/16c3c941409d0581286eff49b180930f-Paper-Conference.pdf},
 volume = {2024},
 year = {2024}
}

@ARTICLE{yu2025metaearth,
  author={Yu, Zhiping and Liu, Chenyang and Liu, Liqin and Shi, Zhenwei and Zou, Zhengxia},
  journal={IEEE Transactions on Pattern Analysis and Machine Intelligence}, 
  title={MetaEarth: A Generative Foundation Model for Global-Scale Remote Sensing Image Generation}, 
  year={2025},
  volume={47},
  number={3},
  pages={1764-1781},
  doi={10.1109/TPAMI.2024.3507010}}

@ARTICLE{liu2025text2earth,
  author={Liu, Chenyang and Chen, Keyan and Zhao, Rui and Zou, Zhengxia and Shi, Zhenwei},
  journal={IEEE Geoscience and Remote Sensing Magazine}, 
  title={Text2Earth: Unlocking text-driven remote sensing image generation with a global-scale dataset and a foundation model}, 
  year={2025},
  volume={13},
  number={3},
  pages={238-259},
  doi={10.1109/MGRS.2025.3560455}}

@inproceedings{
hou2026plan,
title={Plan, Decouple, Assimilate: Physics-Aware Object Insertion in Remote Sensing Imagery},
author={Yingyan Hou and Xianchi Dong and Chao Ren and Wanxuan Lu and Zihan Wei and Hongfeng Yu and Yixiao Wang and Xian Sun},
booktitle={Forty-third International Conference on Machine Learning},
year={2026},
url={https://openreview.net/forum?id=ojx0CyGXHJ}
}

@article{rekavandi2025guide,
  title={A guide to image-and video-based small object detection using deep learning: case study of maritime surveillance},
  author={Rekavandi, Aref Miri and Xu, Lian and Boussaid, Farid and others},
  journal={IEEE Transactions on Intelligent Transportation Systems},
  year={2025}
}

@article{zhou2021aircraft,
  title={Aircraft detection for remote sensing images based on deep convolutional neural networks},
  author={Zhou, Liming and Yan, Haoxin and Shan, Yingzi and others},
  journal={Journal of Electrical and Computer Engineering},
  year={2021}
}

@inproceedings{bandarupally2020detection,
  title={Detection of military targets from satellite images using deep convolutional neural networks},
  author={Bandarupally, Harika and Talusani, Harshitha Reddy and Sridevi, T},
  booktitle={2020 IEEE 5th international conference on computing communication and automation (ICCCA)},
  year={2020}
}

@article{gui2024remote,
AUTHOR = {Gui, Shengxi and Song, Shuang and Qin, Rongjun and Tang, Yang},
TITLE = {Remote Sensing Object Detection in the Deep Learning Era—A Review},
JOURNAL = {Remote Sensing},
VOLUME = {16},
YEAR = {2024},
NUMBER = {2},
ARTICLE-NUMBER = {327},
URL = {https://www.mdpi.com/2072-4292/16/2/327},
ISSN = {2072-4292},
DOI = {10.3390/rs16020327}
}

@article{wang2022remote,
  title={Remote sensing image super-resolution and object detection: Benchmark and state of the art},
  author={Wang, Yi and Bashir, Syed Muhammad Arsalan and Khan, Mahrukh and others},
  journal={Expert Systems with Applications},
  year={2022}
}

@article{gao2024yolo,
  title={Yolo-parallel: Positive gradient modeling for long-tail remote sensing object detection},
  author={Gao, Xiangyi and Zhao, Danpei and Yuan, Zhichao},
  journal={IEEE Geoscience and Remote Sensing Letters},
  year={2024}
}

@inproceedings{chen2024anydoor,
  title={Anydoor: Zero-shot object-level image customization},
  author={Chen, Xi and Huang, Lianghua and Liu, Yu and others},
  booktitle={CVPR},
  year={2024}
}

@inproceedings{goodfellow2014generative,
 author = {Goodfellow, Ian J. and Pouget-Abadie, Jean and Mirza, Mehdi and Xu, Bing and Warde-Farley, David and Ozair, Sherjil and Courville, Aaron and Bengio, Yoshua},
 booktitle = {Advances in Neural Information Processing Systems},
 editor = {Z. Ghahramani and M. Welling and C. Cortes and N. Lawrence and K. Weinberger},
 pages = {},
 publisher = {Curran Associates, Inc.},
 title = {Generative Adversarial Nets},
 url = {https://proceedings.neurips.cc/paper_files/paper/2014/file/f033ed80deb0234979a61f95710dbe25-Paper.pdf},
 volume = {27},
 year = {2014}
}

@article{ho2020denoising,
  title={Denoising diffusion probabilistic models},
  author={Ho, Jonathan and Jain, Ajay and Abbeel, Pieter},
  journal={Advances in neural information processing systems},
  volume={33},
  pages={6840--6851},
  year={2020}
}

@article{dhariwal2021diffusion,
  title={Diffusion models beat gans on image synthesis},
  author={Dhariwal, Prafulla and Nichol, Alexander},
  journal={Advances in neural information processing systems},
  volume={34},
  pages={8780--8794},
  year={2021}
}

@inproceedings{rombach2022high,
  title={High-resolution image synthesis with latent diffusion models},
  author={Rombach, Robin and Blattmann, Andreas and Lorenz, Dominik and Esser, Patrick and Ommer, Bj{\"o}rn},
  booktitle={Proceedings of the IEEE/CVF conference on computer vision and pattern recognition},
  pages={10684--10695},
  year={2022}
}

@inproceedings{hu2022lora,
  title={Lora: Low-rank adaptation of large language models.},
  author={Hu, Edward J and Shen, Yelong and Wallis, Phillip and Allen-Zhu, Zeyuan and Li, Yuanzhi and Wang, Shean and Wang, Lu and Chen, Weizhu and others},
  booktitle={International Conference on Learning Representations (ICLR)},
  year={2022}
}

@inproceedings{tan2025ominicontrol,
  title={Ominicontrol: Minimal and universal control for diffusion transformer},
  author={Tan, Zhenxiong and Liu, Songhua and Yang, Xingyi and Xue, Qiaochu and Wang, Xinchao},
  booktitle={Proceedings of the IEEE/CVF International Conference on Computer Vision},
  pages={14940--14950},
  year={2025}
}

@article{tang2024crs,
  title={Crs-diff: Controllable remote sensing image generation with diffusion model},
  author={Tang, Datao and Cao, Xiangyong and Hou, Xingsong and Jiang, Zhongyuan and Liu, Junmin and Meng, Deyu},
  journal={IEEE Transactions on Geoscience and Remote Sensing},
  year={2024},
  publisher={IEEE}
}

@inproceedings{toker2024satsynth,
  title={Satsynth: Augmenting image-mask pairs through diffusion models for aerial semantic segmentation},
  author={Toker, Aysim and Eisenberger, Marvin and Cremers, Daniel and Leal-Taix{\'e}, Laura},
  booktitle={Proceedings of the IEEE/CVF Conference on Computer Vision and Pattern Recognition},
  pages={27695--27705},
  year={2024}
}

@inproceedings{chen2024zero,
 author = {Chen, Xi and Feng, Yutong and Chen, Mengting and Wang, Yiyang and Zhang, Shilong and Liu, Yu and Shen, Yujun and Zhao, Hengshuang},
 booktitle = {Advances in Neural Information Processing Systems},
 doi = {10.52202/079017-2669},
 editor = {A. Globerson and L. Mackey and D. Belgrave and A. Fan and U. Paquet and J. Tomczak and C. Zhang},
 pages = {84010--84032},
 publisher = {Curran Associates, Inc.},
 title = {Zero-shot Image Editing with Reference Imitation},
 url = {https://proceedings.neurips.cc/paper_files/paper/2024/file/98b2b307aa4aa323df2ba3a83460f25e-Paper-Conference.pdf},
 volume = {37},
 year = {2024}
}

@article{wu2025qwen,
  title={Qwen-image technical report},
  author={Wu, Chenfei and Li, Jiahao and Zhou, Jingren and Lin, Junyang and Gao, Kaiyuan and Yan, Kun and Yin, Sheng-ming and Bai, Shuai and Xu, Xiao and Chen, Yilei and others},
  journal={arXiv preprint arXiv:2508.02324},
  year={2025}
}

@inproceedings{wang2025unicombine,
  title={Unicombine: Unified multi-conditional combination with diffusion transformer},
  author={Wang, Haoxuan and Peng, Jinlong and He, Qingdong and Yang, Hao and Jin, Ying and Wu, Jiafu and Hu, Xiaobin and Pan, Yanjie and Gan, Zhenye and Chi, Mingmin and others},
  booktitle={Proceedings of the IEEE/CVF International Conference on Computer Vision},
  pages={18325--18334},
  year={2025}
}

@inproceedings{song2025insert,
  title={Insert anything: Image insertion via in-context editing in dit},
  author={Song, Wensong and Jiang, Hong and Yang, Zongxin and Cheng, Zheqiao and Quan, Ruijie and Yang, Yi},
  booktitle={Proceedings of the AAAI Conference on Artificial Intelligence},
  volume={40},
  number={11},
  pages={9097--9105},
  year={2026}
}

@inproceedings{zhang2023adding,
  title={Adding conditional control to text-to-image diffusion models},
  author={Zhang, Lvmin and Rao, Anyi and Agrawala, Maneesh},
  booktitle={Proceedings of the IEEE/CVF international conference on computer vision},
  pages={3836--3847},
  year={2023}
}

@article{li2025segearth,
  title={SegEarth-OV3: Exploring SAM 3 for Open-Vocabulary Semantic Segmentation in Remote Sensing Images},
  author={Li, Kaiyu and Zhang, Shengqi and Deng, Yupeng and Wang, Zhi and Meng, Deyu and Cao, Xiangyong},
  journal={arXiv preprint arXiv:2512.08730},
  year={2025}
}

@article{carion2025sam,
  title={Sam 3: Segment anything with concepts},
  author={Carion, Nicolas and Gustafson, Laura and Hu, Yuan-Ting and Debnath, Shoubhik and Hu, Ronghang and Suris, Didac and Ryali, Chaitanya and Alwala, Kalyan Vasudev and Khedr, Haitham and Huang, Andrew and others},
  journal={arXiv preprint arXiv:2511.16719},
  year={2025}
}

@article{liu2024diffusion,
  title={Diffusion models meet remote sensing: Principles, methods, and perspectives},
  author={Liu, Yidan and Yue, Jun and Xia, Shaobo and others},
  journal={IEEE Transactions on Geoscience and Remote Sensing},
  year={2024},
  publisher={IEEE}
}

@inproceedings{lin2014microsoft,
  title={Microsoft coco: Common objects in context},
  author={Lin, Tsung-Yi and Maire, Michael and Belongie, Serge and Hays, James and Perona, Pietro and Ramanan, Deva and Doll{\'a}r, Piotr and Zitnick, C Lawrence},
  booktitle={European conference on computer vision},
  pages={740--755},
  year={2014},
  organization={Springer}
}

@article{liu2022diffusion,
AUTHOR = {Liu, Jinzhe and Yuan, Zhiqiang and Pan, Zhaoying and Fu, Yiqun and Liu, Li and Lu, Bin},
TITLE = {Diffusion Model with Detail Complement for Super-Resolution of Remote Sensing},
JOURNAL = {Remote Sensing},
VOLUME = {14},
YEAR = {2022},
NUMBER = {19},
ARTICLE-NUMBER = {4834},
URL = {https://www.mdpi.com/2072-4292/14/19/4834},
ISSN = {2072-4292},
DOI = {10.3390/rs14194834}
}

@article{wang2025semantic,
  title={Semantic guided large scale factor remote sensing image super-resolution with generative diffusion prior},
  author={Wang, Ce and Sun, Wanjie},
  journal={ISPRS Journal of Photogrammetry and Remote Sensing},
  volume={220},
  pages={125--138},
  year={2025},
  publisher={Elsevier}
}

@article{sousa2025data,
  title={Data augmentation in earth observation: A diffusion model approach},
  author={Sousa, Tiago and Ries, Beno{\^\i}t and Guelfi, Nicolas},
  journal={Information},
  volume={16},
  number={2},
  pages={81},
  year={2025},
  publisher={MDPI}
}

@article{yuan2023efficient,
  title={Efficient and controllable remote sensing fake sample generation based on diffusion model},
  author={Yuan, Zhiqiang and Hao, Chongyang and Zhou, Ruixue and Chen, Jialiang and Yu, Miao and Zhang, Wenkai and Wang, Hongqi and Sun, Xian},
  journal={IEEE Transactions on Geoscience and Remote Sensing},
  volume={61},
  pages={1--12},
  year={2023},
  publisher={IEEE}
}

@article{deng2025synthesizing,
AUTHOR = {Deng, Kai and Wei, Siyuan and Pang, Shiyan and Jiang, Huiwei and Su, Bo},
TITLE = {Synthesizing Remote Sensing Images from Land Cover Annotations via Graph Prior Masked Diffusion},
JOURNAL = {Remote Sensing},
VOLUME = {17},
YEAR = {2025},
NUMBER = {13},
ARTICLE-NUMBER = {2254},
URL = {https://www.mdpi.com/2072-4292/17/13/2254},
ISSN = {2072-4292},
DOI = {10.3390/rs17132254}
}

@article{han2025exploring,
  title={Exploring Text-Guided Single Image Editing for Remote Sensing Images},
  author={Han, Fangzhou and Si, Lingyu and Dong, Hongwei and Jiang, Zhizhuo and Zhang, Lamei and Chen, Hao and Liu, Yu and Du, Bo},
  journal={IEEE Journal of Selected Topics in Applied Earth Observations and Remote Sensing},
  year={2025},
  publisher={IEEE}
}

@article{zhang2025shadow,
  title={Shadow detection and removal for remote sensing images via multi-feature adaptive optimization and geometry-aware illumination compensation},
  author={Zhang, Zhizheng and Cao, Rui and Sheng, Hongting and Guo, Mingqiang and Shao, Zhenfeng and Wu, Liang},
  journal={Expert Systems with Applications},
  pages={127769},
  year={2025},
  publisher={Elsevier}
}

@inproceedings{tsai2017deep,
  title={Deep image harmonization},
  author={Tsai, Yi-Hsuan and Shen, Xiaohui and Lin, Zhe and Sunkavalli, Kalyan and Lu, Xin and Yang, Ming-Hsuan},
  booktitle={Proceedings of the IEEE conference on computer vision and pattern recognition},
  pages={3789--3797},
  year={2017}
}

@inproceedings{cong2020dovenet,
  title={Dovenet: Deep image harmonization via domain verification},
  author={Cong, Wenyan and Zhang, Jianfu and Niu, Li and Liu, Liu and Ling, Zhixin and Li, Weiyuan and Zhang, Liqing},
  booktitle={Proceedings of the IEEE/CVF conference on computer vision and pattern recognition},
  pages={8394--8403},
  year={2020}
}

@article{wang2023samrs,
  title={Samrs: Scaling-up remote sensing segmentation dataset with segment anything model},
  author={Wang, Di and Zhang, Jing and Du, Bo and Xu, Minqiang and Liu, Lin and Tao, Dacheng and Zhang, Liangpei},
  journal={Advances in Neural Information Processing Systems},
  volume={36},
  pages={8815--8827},
  year={2023}
}

@article{sun2022fair1m,
  title={FAIR1M: A benchmark dataset for fine-grained object recognition in high-resolution remote sensing imagery},
  author={Sun, Xian and Wang, Peijin and Yan, Zhiyuan and Xu, Feng and Wang, Ruiping and Diao, Wenhui and Chen, Jin and Li, Jihao and Feng, Yingchao and Xu, Tao and others},
  journal={ISPRS Journal of Photogrammetry and Remote Sensing},
  volume={184},
  pages={116--130},
  year={2022},
  publisher={Elsevier}
}

@inproceedings{waqas2019isaid,
  title={isaid: A large-scale dataset for instance segmentation in aerial images},
  author={Waqas Zamir, Syed and Arora, Aditya and Gupta, Akshita and Khan, Salman and Sun, Guolei and Shahbaz Khan, Fahad and Zhu, Fan and Shao, Ling and Xia, Gui-Song and Bai, Xiang},
  booktitle={Proceedings of the IEEE/CVF conference on computer vision and pattern recognition workshops},
  pages={28--37},
  year={2019}
}

@InProceedings{xia2018dota,
author = {Xia, Gui-Song and Bai, Xiang and Ding, Jian and Zhu, Zhen and Belongie, Serge and Luo, Jiebo and Datcu, Mihai and Pelillo, Marcello and Zhang, Liangpei},
title = {DOTA: A Large-Scale Dataset for Object Detection in Aerial Images},
booktitle = {The IEEE Conference on Computer Vision and Pattern Recognition (CVPR)},
month = {June},
year = {2018}
}

@inproceedings{yang2020rotated,
  title={Rotated faster R-CNN for oriented object detection in aerial images},
  author={Yang, Sheng and Pei, Ziqiang and Zhou, Feng and Wang, Guoyou},
  booktitle={Proceedings of the 2020 3rd International Conference on Robot Systems and Applications},
  pages={35--39},
  year={2020}
}

@article{wenqi2024mar20,
  title={{MAR20}: A Benchmark for Military Aircraft Recognition in Remote Sensing Images},
  author={Yu, Wenqi and Cheng, Gong and Wang, Meijun and Yao, Yanqing and Xie, Xingxing and Yao, Xiwen and Han, Junwei},
  journal={National Remote Sensing Bulletin},
  volume={27},
  number={12},
  pages={2688--2696},
  year={2023},
  doi={10.11834/jrs.20222139}
}

@inproceedings{xie2021oriented,
  title={Oriented R-CNN for object detection},
  author={Xie, Xingxing and Cheng, Gong and Wang, Jiabao and Yao, Xiwen and Han, Junwei},
  booktitle={Proceedings of the IEEE/CVF international conference on computer vision},
  pages={3520--3529},
  year={2021}
}

@article{sapkota2025yolo26,
  title={YOLO26: key architectural enhancements and performance benchmarking for real-time object detection},
  author={Sapkota, Ranjan and Cheppally, Rahul Harsha and Sharda, Ajay and Karkee, Manoj},
  journal={arXiv preprint arXiv:2509.25164},
  year={2025}
}

@article{han2021align,
  title={Align deep features for oriented object detection},
  author={Han, Jiaming and Ding, Jian and Li, Jie and Xia, Gui-Song},
  journal={IEEE transactions on geoscience and remote sensing},
  volume={60},
  pages={1--11},
  year={2021},
  publisher={IEEE}
}

@InProceedings{mao2025ace++,
    author    = {Mao, Chaojie and Zhang, Jingfeng and Pan, Yulin and Jiang, Zeyinzi and Han, Zhen and Liu, Yu and Zhou, Jingren},
    title     = {ACE++: Instruction-Based Image Creation and Editing via Context-Aware Content Filling},
    booktitle = {Proceedings of the IEEE/CVF International Conference on Computer Vision (ICCV) Workshops},
    month     = {October},
    year      = {2025},
    pages     = {1979-1987}
}

@inproceedings{yu2025omnipaint,
  title={Omnipaint: Mastering object-oriented editing via disentangled insertion-removal inpainting},
  author={Yu, Yongsheng and Zeng, Ziyun and Zheng, Haitian and Luo, Jiebo},
  booktitle={Proceedings of the IEEE/CVF International Conference on Computer Vision},
  pages={17324--17334},
  year={2025}
}

@article{burgert2025label,
  title={A label propagation strategy for cutmix in multi-label remote sensing image classification},
  author={Burgert, Tom and Clasen, Kai Norman and Klotz, Jonas and Siebert, Tim and Demir, Beg{\"u}m},
  journal={IEEE Journal of Selected Topics in Applied Earth Observations and Remote Sensing},
  year={2025},
  publisher={IEEE}
}

@article{wei2020hrsid,
  title={HRSID: A high-resolution SAR images dataset for ship detection and instance segmentation},
  author={Wei, Shunjun and Zeng, Xiangfeng and Qu, Qizhe and Wang, Mou and Su, Hao and Shi, Jun},
  journal={Ieee Access},
  volume={8},
  pages={120234--120254},
  year={2020},
  publisher={IEEE}
}

@Article{zhang2021sar,
AUTHOR = {Zhang, Tianwen and Zhang, Xiaoling and Li, Jianwei and Xu, Xiaowo and Wang, Baoyou and Zhan, Xu and Xu, Yanqin and Ke, Xiao and Zeng, Tianjiao and Su, Hao and Ahmad, Israr and Pan, Dece and Liu, Chang and Zhou, Yue and Shi, Jun and Wei, Shunjun},
TITLE = {{SAR} Ship Detection Dataset {(SSDD)}: Official Release and Comprehensive Data Analysis},
JOURNAL = {Remote Sensing},
VOLUME = {13},
YEAR = {2021},
NUMBER = {18},
ARTICLE-NUMBER = {3690},
URL = {https://www.mdpi.com/2072-4292/13/18/3690},
ISSN = {2072-4292},
DOI = {10.3390/rs13183690}
}

@article{li2022mcanet,
  title={MCANet: A joint semantic segmentation framework of optical and SAR images for land use classification},
  author={Li, Xue and Zhang, Guo and Cui, Hao and Hou, Shasha and Wang, Shunyao and Li, Xin and Chen, Yujia and Li, Zhijiang and Zhang, Li},
  journal={International Journal of Applied Earth Observation and Geoinformation},
  volume={106},
  pages={102638},
  year={2022},
  publisher={Elsevier}
}

@article{li2020object,
  title={Object detection in optical remote sensing images: A survey and a new benchmark},
  author={Li, Ke and Wan, Gang and Cheng, Gong and Meng, Liqiu and Han, Junwei},
  journal={ISPRS journal of photogrammetry and remote sensing},
  volume={159},
  pages={296--307},
  year={2020},
  publisher={Elsevier}
}

@article{ding2021object,
  title={Object detection in aerial images: A large-scale benchmark and challenges},
  author={Ding, Jian and Xue, Nan and Xia, Gui-Song and Bai, Xiang and Yang, Wen and Yang, Michael Ying and Belongie, Serge and Luo, Jiebo and Datcu, Mihai and Pelillo, Marcello and others},
  journal={IEEE transactions on pattern analysis and machine intelligence},
  volume={44},
  number={11},
  pages={7778--7796},
  year={2021},
  publisher={IEEE}
}

@article{pan2025earthsynth,
  title={Earthsynth: Generating informative earth observation with diffusion models},
  author={Pan, Jiancheng and Lei, Shiye and Fu, Yuqian and Li, Jiahao and Liu, Yanxing and Sun, Yuze and He, Xiao and Peng, Long and Huang, Xiaomeng and Zhao, Bo},
  journal={arXiv preprint arXiv:2505.12108},
  year={2025}
}

@article{tang2025terragen,
  title={Terragen: A unified multi-task layout generation framework for remote sensing data augmentation},
  author={Tang, Datao and Wang, Hao and Xin, Yudeng and Qiao, Hui and Jiang, Dongsheng and Li, Yin and Yu, Zhiheng and Cao, Xiangyong},
  journal={arXiv preprint arXiv:2510.21391},
  year={2025}
}

@article{yang2025task,
  title={Task-Oriented Data Synthesis and Control-Rectify Sampling for Remote Sensing Semantic Segmentation},
  author={Yang, Yunkai and Zhang, Yudong and Zhang, Kunquan and Zhang, Jinxiao and Chen, Xinying and Fu, Haohuan and Dong, Runmin},
  journal={arXiv preprint arXiv:2512.16740},
  year={2025}
}

@inproceedings{fu2025univg,
  title={Univg: A generalist diffusion model for unified image generation and editing},
  author={Fu, Tsu-Jui and Qian, Yusu and Chen, Chen and Hu, Wenze and Gan, Zhe and Yang, Yinfei},
  booktitle={Proceedings of the IEEE/CVF International Conference on Computer Vision},
  pages={17160--17170},
  year={2025}
}

@inproceedings{yu2025anyedit,
  title={Anyedit: Mastering unified high-quality image editing for any idea},
  author={Yu, Qifan and Chow, Wei and Yue, Zhongqi and Pan, Kaihang and Wu, Yang and Wan, Xiaoyang and Li, Juncheng and Tang, Siliang and Zhang, Hanwang and Zhuang, Yueting},
  booktitle={Proceedings of the Computer Vision and Pattern Recognition Conference},
  pages={26125--26135},
  year={2025}
}

@article{niu2021making,
  title={Making images real again: A comprehensive survey on deep image composition},
  author={Niu, Li and Cong, Wenyan and Liu, Liu and Hong, Yan and Zhang, Bo and Liang, Jing and Zhang, Liqing},
  journal={arXiv preprint arXiv:2106.14490},
  year={2021}
}

@inproceedings{zhang2025zerocomp,
  title={Zerocomp: Zero-shot object compositing from image intrinsics via diffusion},
  author={Zhang, Zitian and Fortier-Chouinard, Fr{\'e}d{\'e}ric and Garon, Mathieu and Bhattad, Anand and Lalonde, Jean-Fran{\c{c}}ois},
  booktitle={2025 IEEE/CVF Winter Conference on Applications of Computer Vision (WACV)},
  pages={483--494},
  year={2025},
  organization={IEEE}
}

@article{schouten2026hiddenobjects,
  title={HiddenObjects: Scalable Diffusion-Distilled Spatial Priors for Object Placement},
  author={Schouten, Marco and Siglidis, Ioannis and Belongie, Serge and Papadopoulos, Dim P},
  journal={arXiv preprint arXiv:2604.10675},
  year={2026}
}

@article{zhang2025region,
  title={Region-to-Region: Enhancing Generative Image Harmonization with Adaptive Regional Injection},
  author={Zhang, Zhiqiu and Fan, Dongqi and Wang, Mingjie and Tang, Qiang and Yang, Jian and Yi, Zili},
  journal={arXiv preprint arXiv:2508.09746},
  year={2025}
}

@inproceedings{
fortier2024spotlight,
title={SpotLight: Shadow-Guided Object Relighting via Diffusion},
author={Fr{\'e}d{\'e}ric Fortier-Chouinard and Zitian Zhang and Louis-Etienne Messier and Mathieu Garon and Anand Bhattad and Jean-Francois Lalonde},
booktitle={Thirteenth International Conference on 3D Vision},
year={2026},
url={https://openreview.net/forum?id=mZw9TOQUUd}
}

@inproceedings{li2025aicomposer,
  title={AIComposer: Any Style and Content Image Composition via Feature Integration},
  author={Li, Haowen and Fan, Zhenfeng and Wen, Zhang and Zhu, Zhengzhou and Li, Yunjin},
  booktitle={Proceedings of the IEEE/CVF International Conference on Computer Vision},
  pages={16840--16850},
  year={2025}
}

@inproceedings{zhang2025freeinsert,
  title={FreeInsert: Personalized Object Insertion with Geometric and Style Control},
  author={Zhang, Yuhong and Wang, Han and Wang, Yiwen and Xie, Rong and Song, Li},
  booktitle={Proceedings of the 33rd ACM International Conference on Multimedia},
  pages={10361--10369},
  year={2025}
}

@inproceedings{li2025freeinsert,
  title={Freeinsert: Disentangled text-guided object insertion in 3d gaussian scene without spatial priors},
  author={Li, Chenxi and Wang, Weijie and Li, Qiang and Sebe, Nicu and Lepri, Bruno and Nie, Weizhi},
  booktitle={Proceedings of the 33rd ACM International Conference on Multimedia},
  pages={10915--10924},
  year={2025}
}

@inproceedings{sheng2021ssn,
  title={Ssn: Soft shadow network for image compositing},
  author={Sheng, Yichen and Zhang, Jianming and Benes, Bedrich},
  booktitle={Proceedings of the IEEE/CVF Conference on Computer Vision and Pattern Recognition},
  pages={4380--4390},
  year={2021}
}

@ARTICLE{zhenyuan2026rsedit,
  author={Chen, Zhenyuan and Zhang, Zechuan and Zhang, Feng},
  journal={IEEE Geoscience and Remote Sensing Letters}, 
  title={RSEdit: Text-Guided Image Editing for Remote Sensing}, 
  year={2026},
  volume={23},
  number={},
  pages={6011905-6011905},
  doi={10.1109/LGRS.2026.3695484}}

@article{tang2024changeanywhere,
  title={Changeanywhere: Sample generation for remote sensing change detection via semantic latent diffusion model},
  author={Tang, Kai and Chen, Jin},
  journal={arXiv preprint arXiv:2404.08892},
  year={2024}
}

@article{zhang2026geodiff,
  title={GeoDiff-SAR: A Geometric Prior Guided Diffusion Model for SAR Image Generation},
  author={Zhang, Fan and Wu, Xuanting and Ma, Fei and Yin, Qiang and Hu, Yuxin},
  journal={arXiv preprint arXiv:2601.03499},
  year={2026}
}

@article{gong2025crossearth,
  author={Gong, Ziyang and Wei, Zhixiang and Wang, Di and Hu, Xiaoxing and Ma, Xianzheng and Chen, Hongruixuan and Jia, Yuru and Deng, Yupeng and Ji, Zhenming and Zhu, Xiangwei and Yang, Xue and Yokoya, Naoto and Zhang, Jing and Du, Bo and Yan, Junchi and Zhang, Liangpei},
  journal={IEEE Transactions on Pattern Analysis and Machine Intelligence}, 
  title={CrossEarth: Geospatial Vision Foundation Model for Domain Generalizable Remote Sensing Semantic Segmentation}, 
  year={2026},
  volume={48},
  number={5},
  pages={5147-5164},
  doi={10.1109/TPAMI.2025.3649001}}

@article{li2025saratr,
  title={SARATR-X: Toward building a foundation model for SAR target recognition},
  author={Li, Weijie and Yang, Wei and Hou, Yuenan and Liu, Li and Liu, Yongxiang and Li, Xiang},
  journal={IEEE Transactions on Image Processing},
  volume={34},
  pages={869--884},
  year={2025},
  publisher={IEEE}
}

@article{debuysere2025quantitative,
  title={Quantitative comparison of fine-tuning techniques for pretrained latent diffusion models in the generation of unseen SAR images},
  author={Debuys{\`e}re, Sol{\`e}ne and Trouv{\'e}, Nicolas and Letheule, Nathan and L{\'e}v{\^e}que, Olivier and Colin, Elise},
  journal={ISPRS Journal of Photogrammetry and Remote Sensing},
  volume={234},
  pages={93--110},
  year={2026},
  publisher={Elsevier}
}

@article{martin2026generating,
  title={Generating Satellite Imagery Data for Wildfire Detection through Mask-Conditioned Generative AI},
  author={Martin, Valeria and Venable, K Brent and Morgan, Derek},
  journal={arXiv preprint arXiv:2604.02479},
  year={2026}
}

@inproceedings{phongthawee2024diffusionlight,
  title={Diffusionlight: Light probes for free by painting a chrome ball},
  author={Phongthawee, Pakkapon and Chinchuthakun, Worameth and Sinsunthithet, Nontaphat and Jampani, Varun and Raj, Amit and Khungurn, Pramook and Suwajanakorn, Supasorn},
  booktitle={Proceedings of the IEEE/CVF conference on computer vision and pattern recognition},
  pages={98--108},
  year={2024}
}

@inproceedings{ke2024repurposing,
  title={Repurposing diffusion-based image generators for monocular depth estimation},
  author={Ke, Bingxin and Obukhov, Anton and Huang, Shengyu and Metzger, Nando and Daudt, Rodrigo Caye and Schindler, Konrad},
  booktitle={Proceedings of the IEEE/CVF conference on computer vision and pattern recognition},
  pages={9492--9502},
  year={2024}
}

@article{lipman2022flow,
  title={Flow matching for generative modeling},
  author={Lipman, Yaron and Chen, Ricky TQ and Ben-Hamu, Heli and Nickel, Maximilian and Le, Matt},
  journal={arXiv preprint arXiv:2210.02747},
  year={2022}
}

@inproceedings{zeng2024rgbx,
author = {Zeng, Zheng and Deschaintre, Valentin and Georgiev, Iliyan and Hold-Geoffroy, Yannick and Hu, Yiwei and Luan, Fujun and Yan, Ling-Qi and Ha\v{s}an, Milo\v{s}},
title = {RGB$\leftrightarrow$X: Image decomposition and synthesis using material- and lighting-aware diffusion models},
year = {2024},
isbn = {9798400705250},
publisher = {Association for Computing Machinery},
doi = {10.1145/3641519.3657445},
booktitle = {ACM SIGGRAPH 2024 Conference Papers},
articleno = {75},
numpages = {11},
series = {SIGGRAPH '24}
}

@ARTICLE{finegrip,
  author={Zhao, Danpei and Yuan, Bo and Chen, Ziqiang and Li, Tian and Liu, Zhuoran and Li, Wentao and Gao, Yue},
  journal={IEEE Transactions on Geoscience and Remote Sensing}, 
  title={Panoptic Perception: A Novel Task and Fine-Grained Dataset for Universal Remote Sensing Image Interpretation}, 
  year={2024},
  volume={62},
  number={},
  pages={1-14},
  doi={10.1109/TGRS.2024.3392778}}

@article{yu2021bisenet,
  title={Bisenet v2: Bilateral network with guided aggregation for real-time semantic segmentation},
  author={Yu, Changqian and Gao, Changxin and Wang, Jingbo and Yu, Gang and Shen, Chunhua and Sang, Nong},
  journal={International journal of computer vision},
  volume={129},
  number={11},
  pages={3051--3068},
  year={2021},
  publisher={Springer}
}

@inproceedings{xu2023pidnet,
  title={PIDNet: A real-time semantic segmentation network inspired by PID controllers},
  author={Xu, Jiacong and Xiong, Zixiang and Bhattacharyya, Shankar P},
  booktitle={2023 IEEE/CVF Conference on Computer Vision and Pattern Recognition (CVPR)},
  pages={19529--19539},
  year={2023},
  organization={IEEE}
}

@article{guo2022segnext,
  title={Segnext: Rethinking convolutional attention design for semantic segmentation},
  author={Guo, Meng-Hao and Lu, Cheng-Ze and Hou, Qibin and Liu, Zhengning and Cheng, Ming-Ming and Hu, Shi-Min},
  journal={Advances in neural information processing systems},
  volume={35},
  pages={1140--1156},
  year={2022}
}

@article{xie2021segformer,
  title={SegFormer: Simple and efficient design for semantic segmentation with transformers},
  author={Xie, Enze and Wang, Wenhai and Yu, Zhiding and Anandkumar, Anima and Alvarez, Jose M and Luo, Ping},
  journal={Advances in neural information processing systems},
  volume={34},
  pages={12077--12090},
  year={2021}
}

\end{document}